%% file: main_arxiv.tex
\documentclass[sigconf]{acmart}

\AtBeginDocument{%
  }

\input{commands}

\setcopyright{none}
\copyrightyear{XXXX}
\acmYear{XXXX}
\acmDOI{XXXXXXX.XXXXXXX}
\acmConference[XXXX]{XXXX}{XXXX}{XXXX}
\acmISBN{XXXX}

\usepackage{tabularx}
\usepackage{multirow}
\usepackage{subfigure}
\usepackage{arydshln}
\usepackage{algorithm}
\usepackage{algorithmic}
\usepackage{enumitem}

\begin{document}

\title{LakeHopper: Knowledge-Aware Adaptation of Column Type Annotators across Data Lakes}

\author{Yushi Sun}
\affiliation{%
  \institution{HKUST}
  \city{Hong Kong}
  \country{China}}
\email{ysunbp@connect.ust.hk}

\author{Xujia Li}
\authornote{Corresponding author.}
\affiliation{%
  \institution{HKUST}
  \city{Hong Kong}
  \country{China}}
\email{leexujia@ust.hk}

\author{Nan Tang}
\affiliation{%
  \institution{HKUST(GZ)}
  \city{Guangzhou}
  \country{China}}
\email{nantang@hkust-gz.edu.cn}

\author{Quanqing Xu}
\affiliation{%
  \institution{Ant Group}
  \city{Hangzhou}
  \country{China}}
\email{xuquanqing.xqq@oceanbase.com}

\author{Chuanhui Yang}
\affiliation{%
  \institution{Ant Group}
  \city{Hangzhou}
  \country{China}}
\email{rizhao.ych@oceanbase.com}

\author{Lei Chen}
\affiliation{%
  \institution{HKUST(GZ)/HKUST}
  \city{Guangzhou/Hong Kong}
  \country{China}}
\email{leichen@cse.ust.hk}

\renewcommand{\shortauthors}{Sun, Li, Tang, Xu, Yang, and Chen}

\begin{abstract}
Column Type Annotation (CTA), which assigns a semantic type (\eg \textit{Film}, \textit{Scientist}) to a table column, underpins data integration, cleaning, and search over data lakes. State-of-the-art annotators are pre-trained language models (PLMs) fine-tuned on one particular corpus of tables, \ie a \textit{source} data lake, and they degrade sharply once deployed on a new (\ie \textit{target}) lake, whose tables and semantic type set both differ. Retraining per lake is prohibitive because it demands large volumes of expert annotations.
We recast cross-lake adaptation as a \emph{knowledge management} problem and make the resulting decomposition explicit: relative to a target annotator, a source annotator holds knowledge that must be \emph{discarded} (source-specific), \emph{realigned and reused} (shared), or \emph{acquired} (target-specific). This decomposition exposes which part of the gap a general-purpose LLM can close and which part only target supervision can. Guided by it, we present \sys, which adapts a source annotator under a fixed annotation budget through three coupled mechanisms: label-set realignment that transplants the output layer for shared types, LLM-\emph{verified} gap discovery that localizes columns the annotator handles unreliably, and cluster-based propagation plus rehearsal fine-tuning that generalizes each flagged column into a labeling batch without erasing shared knowledge. Casting the LLM as a \emph{verifier} of the annotator's own predictions rather than an annotator keeps every output inside the target type set, so \sys structurally emits no out-of-domain labels, whereas prompted LLMs hallucinate types on 2.7--47.6\% of columns.
Across three data lake transfers of differing difficulty, \sys lifts three PLM backbones by up to 71.4\% relative macro-F1, reaches near-full-data quality with under 6\% of the target labels, and matches fine-tuned table LLMs while training 27--131$\times$ faster.
\end{abstract}



\maketitle

\input{secs/sec_introduction}
\input{secs/sec_problem}
\input{secs/sec_methodology}
\input{secs/sec_experiment}
\input{secs/sec_limitations}
\input{secs/sec_related}
\input{secs/sec_conclusion}

\section*{Ethical Considerations}
\label{sec:ethics}

\sys assigns semantic types to table columns; it neither decides about individuals nor produces free-form text about them. Four considerations nonetheless deserve mention.

\stitle{Data provenance and licensing.} All corpora are public research datasets: PublicBI (MIT), VizNet (CC BY 4.0), Semtab2019 (Apache 2.0), GitTables (CC BY 4.0). We used them within their licenses for research only. Because they are scraped from the web and public dashboards, cells may incidentally contain personal information (\eg names in a \textit{Person} column); we neither attempt re-identification nor release record-level derived data.

\stitle{Third-party LLM services.} Gap discovery transmits column values to a hosted LLM API, which in a deployment over sensitive or regulated tables constitutes disclosure to a third party. Practitioners should treat it accordingly --- restricting the queried subset, redacting values, or substituting a local model. Our robustness result (\S\ref{sec:add-robust}) shows the mechanism works with an 8B open-weight verifier runnable on premises, so the design does not depend on an external provider.

\stitle{Failure modes in deployment.} Types produced by \sys feed downstream preparation, so systematic errors propagate. Accuracy in the low-budget regime, while much improved over baselines, remains well below the full-supervision ceiling; \sys is therefore a means to reduce annotation effort with a human in the loop, not an unsupervised replacement for curation. We report residual error levels openly, not only relative gains, so practitioners can judge fitness for their setting.

\stitle{Environmental cost.} Adaptation runs are short --- tens of GPU-minutes plus roughly \textdollar$0.6$--\textdollar$2.6$ of API usage per transfer --- markedly cheaper than the fine-tuned-LLM alternative. Reducing the compute needed to move an annotator to a new lake is one of this work's practical benefits.

\stitle{Reproducibility.} The paper is self-contained: every claim rests on a number reported in \S\ref{sec:experiment}, and \S\ref{sec:experiment} states the settings needed to interpret them. The appendix adds the verbatim verification template, pseudocode with the full complexity derivation, per-dataset statistics and type-set overlaps, per-baseline descriptions, every hyperparameter value, and the tables behind the analyses \S\ref{sec:analyses} reports numerically. All four corpora are public. We will release the adaptation code and per-transfer scripts upon publication.

\clearpage

\appendix
\input{supplementary_arxiv}

\bibliographystyle{ACM-Reference-Format}
\bibliography{main}

\end{document}

%% file: commands.tex
\newcommand{\eg}{\textit{e.g.,\,}}
\newtheorem{example}{Example}
\newcommand{\eop}{\hspace*{\fill}\mbox{$\Box$}}
\newcommand{\be}{\begin{enumerate}}
\newcommand{\ee}{\end{enumerate}}
\newcommand{\ie}{\textit{i.e.,\,}}
\usepackage{xspace}
\newcommand{\sys}{LakeHopper\xspace}

\newcommand{\stab}{\vspace{1.2ex}\noindent}
\newcommand{\sstab}{\rule{0pt}{8pt}\\[-2.2ex]}
\newcommand{\stitle}[1]{\stab\noindent{\bf #1}}
\newcommand{\bi}{\begin{itemize}}
\newcommand{\ei}{\end{itemize}}

\NewDocumentCommand{\yushi}{ mO{} }{\textcolor{red}{\textbf{\small#1}}}

%% file: secs/sec_introduction.tex
\section{Introduction}
\label{sec:introduction}

Column Type Annotation (CTA) labels the semantic data type (\eg \textit{Film}, \textit{Scientist}) of a column in a data lake. It underpins data mining~\cite{sun2023reca}, integration~\cite{hulsebos2019sherlock, rahm2001survey, venetis2011recovering, zhang2019sato}, and cleaning~\cite{raman2001potter, kandel2011wrangler}: knowing a column holds universities rather than arbitrary strings is what lets a system join, validate, or index it meaningfully.

State-of-the-art annotators are \emph{pre-trained language models} (PLMs) --- transformer encoders such as BERT~\cite{devlin2019bert}, pre-trained on text then fine-tuned with a classification layer on top. In CTA the encoder maps a column's serialized cell values to a vector, and the classification layer maps that vector to one type in a fixed type set~\cite{suhara2022annotating, sun2023reca, iida2021tabbie, wang2022sudowoodo}. This is exactly why such annotators are accurate yet immobile: both the fine-tuned weights and the \emph{shape} of the classification layer are tied to their training lake, so moving one to a new lake means paying for a new round of expert annotation. That cost is the bottleneck we attack.

\begin{figure}[t!]
  \centering
\includegraphics[width=1.0\linewidth]{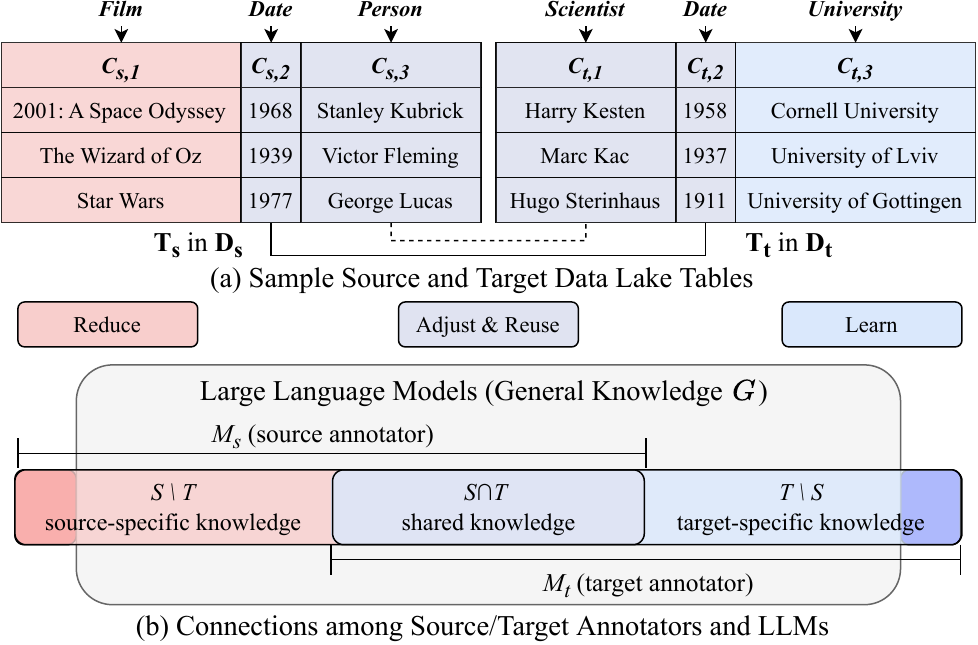}
  \vspace{-1.7em}
  \caption{(a) A cross data lake CTA example. (b) Knowledge of fine-tuned annotators ($S$, $T$ for source and target) and of a generic LLM ($G$).}
  \label{fig:venn}
  \vspace{-1.4em}
\end{figure}

\stitle{Existing Solutions and Limitations.}
Applying an annotator \textit{directly} to an unseen lake fails, because lakes differ both in table content and in which types they distinguish. \citet{langenecker2023steered} report a Sato~\cite{zhang2019sato} annotator dropping from $90\%$ to $35\%$ accuracy when transferred to PublicBI~\cite{vogelsgesang2018get}. Example~\ref{exam:cross} shows why.

\begin{example}
\label{exam:cross}
Figure~\ref{fig:venn}(a) shows $T_s=(C_{s,1}, C_{s,2}, C_{s,3})$ from a {\em source} film lake $D_s$ and $T_t=(C_{t,1}, C_{t,2}, C_{t,3})$ from a {\em target} mathematician lake $D_t$. Let $M_s$ be the source annotator. It cannot label $C_{t,3}$ (\textit{\textbf{University}}) at all: that type is absent from its classification layer, so the correct answer is not even expressible. On familiar content it still errs --- $C_{t,1}$ needs \textit{\textbf{Scientist}} where $M_s$ learned the coarser \textit{\textbf{Person}}. And $C_{s,1}$ (\textit{\textbf{Film}}) is a type $D_t$ never uses, so keeping it competitive can only hurt. Even genuinely shared knowledge must therefore be \emph{realigned} before reuse in $D_t$. \eop
\end{example}

The {\bf central limitation} is that these failures have different causes, so no single remedy fixes them. Collecting a large annotated set on $D_t$ and retraining fixes all three, but is precisely the cost we want to avoid. One might hope an LLM sidesteps the issue, needing no per-lake training. It does not: prompted LLMs underperform trained PLM-based annotators, lacking the domain- and task-specific knowledge a lake encodes (\ie $(T \setminus S) \setminus G$ in Figure~\ref{fig:venn}(b)), and they \emph{hallucinate types outside the requested type set} on a substantial fraction of columns (\S\ref{sec:no-tune}). An annotation absent from the schema is not a usable annotation.

\begin{figure}[t!]
  \centering
  \includegraphics[width=1.0\linewidth]{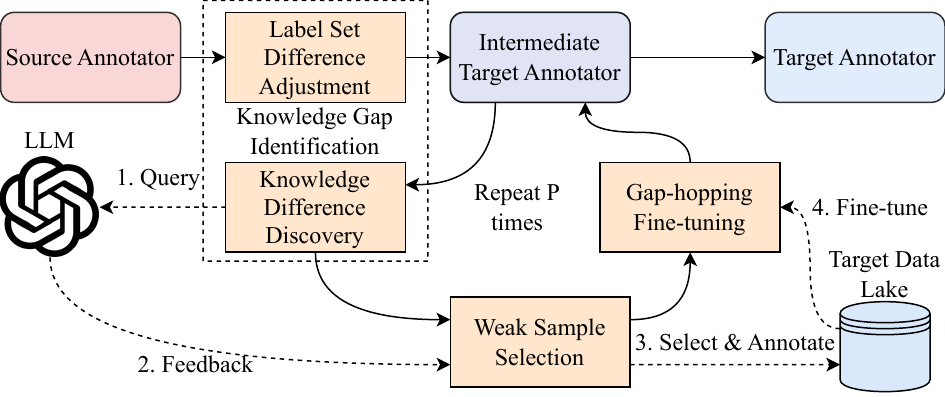}
    \vspace{-1.7em}
  \caption{Architecture of \sys. Steps 1--2 realign the output layer for shared types (\S\ref{sec:label-diff}) and localize unreliable columns via softmax confidence plus LLM verification (\S\ref{sec:knowledge-diff}); step 3 propagates each flagged column to its $K$-means neighborhood and requests a labeled batch (\S\ref{sec:weak-sample}); step 4 fine-tunes with rehearsal over all batches so far (\S\ref{sec:gap-hopping-finetuning}).}
  \label{fig:pipeline}
  \vspace{-1.4em}
\end{figure}

\stitle{Opportunities.}
Trained annotators already exist for many lakes, so the question is not how to build one but how to \emph{move} one --- fundamentally a \emph{knowledge management} problem. Retraining discards knowledge still valid; applying $M_s$ unchanged retains knowledge that is not. A principled adaptation treats the source annotator's knowledge as three separable parts, so obtaining $M_{t}$ for $D_t$ under minimal supervision requires (Figure~\ref{fig:venn}(b)):
{\bf i) discarding} source-specific knowledge ($S \setminus T$, \eg \textit{Film} for $C_{s,1}$);
{\bf ii) realigning and reusing} shared knowledge ($S \cap T$, \eg the \textit{Person}/\textit{Scientist} relation between $C_{s,3}$ and $C_{t,1}$); and
{\bf iii) acquiring} target-specific knowledge ($T \setminus S$, \eg \textit{University} for $C_{t,3}$).

This decomposition makes the LLM's role precise rather than opportunistic. Intersecting it with a generic model's knowledge $G$ splits what must be acquired: $(T\setminus S)\cap G$ is what an LLM can help localize, while $(T\setminus S)\setminus G$ is lake-specific and reachable only through target supervision. The three needs yield three opportunities --- {\bf1)} identify the gap, \ie which target columns $M_s$ handles poorly and which shared knowledge needs realignment; {\bf2)} select minimal yet informative training data; {\bf3)} fine-tune without forgetting --- none of them trivial. The gap resists location from either side: an annotator can be strong on a type in $D_s$ and weak on it in $D_t$ because cell values differ~\cite{langenecker2021towards}, while a PLM is a black box whose obvious proxy, its own confidence, is least reliable exactly under shift, where classifiers are overconfident~\cite{guo2017calibration}. Turning flagged columns into a batch that repays its cost is unexplored in CTA. And fine-tuning must drop, add, and retain types at once, while the model's weak points move after every update.

\begin{table}[t!]
\caption{Comparing existing methods and \sys. \emph{OOD} is the fraction of columns given a label outside $S_t$ (\S\ref{sec:problem-cross}); such labels are unusable downstream regardless of accuracy.}
\vspace{-1.2em}
\centering
\label{tab:comparison}
\small
\begin{tabular}{llll}
\hline
                           & \textbf{Generalizability} & \textbf{Accuracy}  & \textbf{OOD rate} \\ \hline
\textbf{PLM-based}  & low                       & high              & zero                   \\
\textbf{LLM-based*} & high                      & low                     & high                   \\
\textbf{LLM-based+} & low                       & high                           & low                    \\
\textbf{\sys}        & \textbf{high}             & \textbf{high}             & \textbf{zero}     \\
\hline
\multicolumn{4}{l}{* and + are LLMs without or with domain-specific fine-tuning.}\\
\end{tabular}
\vspace{-1.6em}
\end{table}

\stitle{Our Proposal.}
Figure~\ref{fig:pipeline} overviews \sys, whose steps map one-to-one onto these challenges. \sys realigns the source output layer to the target type set, transplanting rows of shared types and re-initializing the rest, then warms up on a small random sample. It runs the annotator over target columns and \emph{verifies} its predictions with an LLM (step 1), flagging columns where the annotator is unreliable, \ie $(T\setminus S)\cap G$ plus shared knowledge invalidated by type-set mismatch ({\bf Challenge 1}). Since querying every column is uneconomical and a few flagged columns make too small a batch, we cluster the lake in the annotator's embedding space and sample a batch from clusters containing a flagged column (steps 2--3) ({\bf Challenge 2}). Fine-tuning then proceeds with rehearsal over all batches so far (step 4), absorbing $T \setminus S$ and letting $S \setminus T$ decay without discarding $T \cap S$ ({\bf Challenge 3}). Because weak points move as the annotator improves, steps 1--4 iterate, with early stopping once the LLM stops contributing.

One choice deserves emphasis, reconciling the two roles we give the LLM: we argued LLMs are poor \emph{annotators}, yet use one as a \emph{verifier}. These differ. Verification is strictly easier --- a binary judgment on a proposed label, not a choice among 78 or 275 types --- and, critically, the LLM never emits a label: predictions come only from the PLM's output layer, defined over $S_t$ by construction. \sys thus inherits an LLM's breadth for \emph{deciding where to look} while structurally guaranteeing zero out-of-domain output (Table~\ref{tab:comparison}). \S\ref{sec:why-verifier} makes this precise, \S\ref{sec:reliability-veri} tests it.

Both existing families give up something \sys keeps (Table~\ref{tab:comparison}). PLM-based approaches~\cite{suhara2022annotating, sun2023reca, iida2021tabbie, wang2022sudowoodo, fan2024unicorn} are accurate and never produce out-of-schema labels but generalize poorly and demand adaptation data; LLM-based approaches~\cite{korini2023column, zhang2024tablellama, li2024table, feuer2023archetype} generalize but, unfine-tuned, annotate poorly and hallucinate out-of-schema types, while fine-tuning repairs accuracy at a compute cost forfeiting the portability that motivated them.

\stitle{Contributions.}

\sstab(1) \textbf{Problem formulation.} We formalize cross-lake CTA and state precisely what is ``crossed'': a label-set shift together with a content shift (\S\ref{sec:problem-cross}). Our \emph{knowledge decomposition} separates the source annotator's knowledge into source-specific, shared, and target-specific parts, intersected with generic model knowledge to delimit what an LLM can supply, turning adaptation from one undifferentiated retraining step into three separately addressable ones. It applies to label-set-shift problems beyond CTA.

\sstab(2) \textbf{LLM-as-verifier adaptation.} We propose \sys, in which the LLM is a domain-agnostic \emph{verifier} localizing the gap, never an annotator. Confining predictions to the target type set \emph{eliminates out-of-domain output by construction}, against the $2.7\%$--$47.6\%$ OOD rates we measure for prompted LLMs, and we argue the trade-off explicitly (\S\ref{sec:why-verifier}).

\sstab(3) \textbf{Budget-aware selection and forgetting-resistant fine-tuning.} We couple cluster-based propagation, turning each flagged column into an informative batch, with rehearsal fine-tuning that preserves shared knowledge. \sys needs only $1.6\%$--$5.9\%$ of target training data for substantial gains and adapts \textbf{27--131$\times$ faster} than a fine-tuned table LLM.

\sstab(4) \textbf{Empirical study across three transfers.} We evaluate transfers spanning pure type-set extension, extension with forgetting, and a distinct corpus, under low and high budgets. \sys improves three backbones consistently, by up to \textbf{71.4\% relative MA F1}, nearing full-data quality with under $6\%$ of labels. We also report where our evidence stops short of our claims (\S\ref{sec:limitations}).

%% file: secs/sec_problem.tex
\section{Problem Definition and Background}
\label{sec:problem}

\subsection{Column Type Annotation}

Let $T$ be a table with columns $\{C_1, \ldots, C_m\}$. A data lake $D$ is a collection of such tables with a type set $S$: a finite set of semantic types from an ontology, flattened to be mutually disjoint and non-hierarchical as practical deployments require. Following standard practice~\cite{zhang2019sato, sun2023reca}, each annotated column carries \emph{exactly one} type from $S$, an assumption inherited from the datasets and annotators we adapt, all treating CTA as single-label multi-class classification.

\stitle{Table Column Type Annotation (CTA).}
Design $f(\cdot)$ mapping a column $C_i$ to a type $\bar{y_i} = f(C_i) \in S$ such that each cell of $C_i$ is an instance of $\bar{y_i}$. As in prior work~\cite{hulsebos2019sherlock, iida2021tabbie}, prediction uses column values only: names and metadata are withheld, being frequently missing, uninformative, or inconsistent across a lake.

\subsection{Cross Data Lake CTA}
\label{sec:problem-cross}

A recurring ambiguity in the transfer literature is what exactly ``crosses'' between corpora, so we state it. Given a source lake $D_s$ with type set $S_s$ ($n_s = |S_s|$) and a target lake $D_t$ with $S_t$ ($n_t = |S_t|$), two shifts occur simultaneously.

\stitle{(i) Label-set shift.} $S_s \neq S_t$ in general, and three regions demand different treatment: $S_s \setminus S_t$, types the target never uses, whose capacity should be released; $S_s \cap S_t$, shared types whose learned boundaries are candidates for reuse; and $S_t \setminus S_s$, types absent from the source classification layer, hence inexpressible by $M_s$ before adaptation. The case $S_s \subset S_t$ is \emph{pure type-set extension}, where nothing must be forgotten. The shift is not purely set-theoretic: two lakes may share a type name while defining it at different granularity, as with \textit{Person} versus \textit{Scientist} in Example~\ref{exam:cross}. Such a type sits in $S_s \cap S_t$ set-wise yet needs its boundary relearned, which is why we speak of \emph{realigning} shared knowledge rather than copying it.

\stitle{(ii) Content shift.} The distribution over column values differs, $P_s(C) \neq P_t(C)$, as lakes differ in provenance, table shape, and surface conventions. A shared type can therefore still be annotated poorly in $D_t$, so label-set alignment alone is insufficient.

\stitle{Cross Data Lakes Column Type Annotation.}
Given $M_s$ fine-tuned on $D_s$, a target lake $D_t$, and a budget $N_t$, select at most $N_t$ samples --- each a $(C_i, y_i)$ pair with $y_i \in S_t$ --- from $D_t$ and use them to obtain $M_t$ maximizing accuracy on $D_t$. The budget is the crux: it models adaptation being limited by affordable expert labeling, which is why we report results as a function of budget rather than at one operating point.

\stitle{Out-of-domain (OOD) output.} A prediction is \emph{out-of-domain} if it falls outside $S_t$; for labels $\{\hat{y}_i\}$ over $N$ columns the OOD rate is $\frac{1}{N}|\{i : \hat{y}_i \notin S_t\}|$. This concerns the output space, not correctness: an OOD label is unusable by a pipeline expecting schema $S_t$, however plausible. Classifier-based annotators have zero OOD rate by construction; generative LLMs do not, so we track it separately.

\stitle{Scope.} We assume one given source annotator, a flat single-label type set, and a target labeling process (human, heuristic, or knowledge-base-backed) invocable up to $N_t$ times. Selecting or combining multiple sources, and hierarchical or multi-label type sets, lie outside this scope; see Section~\ref{sec:limitations}.

\subsection{Background: PLM-based Annotators}
\label{sec:background-plm}

Since our method intervenes on annotator architecture, we recall its two parts. A PLM-based annotator is a transformer \emph{encoder} plus a linear \emph{output layer}. The encoder serializes a column's cell values into an $m$-dimensional representation ($m=768$ for BERT-base); backbones differ in encoded context --- all table columns jointly~\cite{suhara2022annotating}, contrastive column representations~\cite{wang2022sudowoodo}, related-table augmentation~\cite{sun2023reca} --- but all expose this interface. The output layer is $L \in \mathbb{R}^{m \times n}$ whose $j$-th column scores type $s_j$, a softmax over $Lv$ giving the predicted distribution.

Two consequences drive our design. First, $n = |S|$ is baked into $L$'s shape, so an annotator cannot express a type absent from its training lake --- adaptation must modify $L$ first. Second, because $L$ is indexed by type, its columns can be manipulated \emph{per type}: those for shared types transplanted, others re-initialized. \S\ref{sec:label-diff} exploits this.

%% file: secs/sec_methodology.tex
\section{The \sys Framework}
\label{sec:methodology}

\sys adapts $M_s$ into $M_t$ under budget $N_t$ via Figure~\ref{fig:pipeline}'s four steps: \S\ref{sec:label-diff} realigns the output layer as a one-off preparation, and \S\S\ref{sec:knowledge-diff}--\ref{sec:gap-hopping-finetuning} form the iteration --- locate weak points, turn them into a labeling batch, fine-tune, repeat. Write $\Bar{M}_{t,l}$ for the intermediate annotator after iteration $l$.

\subsection{Knowledge Gap Identification}
\subsubsection{Label Set Difference Adjustment} \label{sec:label-diff}

\begin{figure}[t!]
  \centering
  \includegraphics[width=0.92\linewidth]{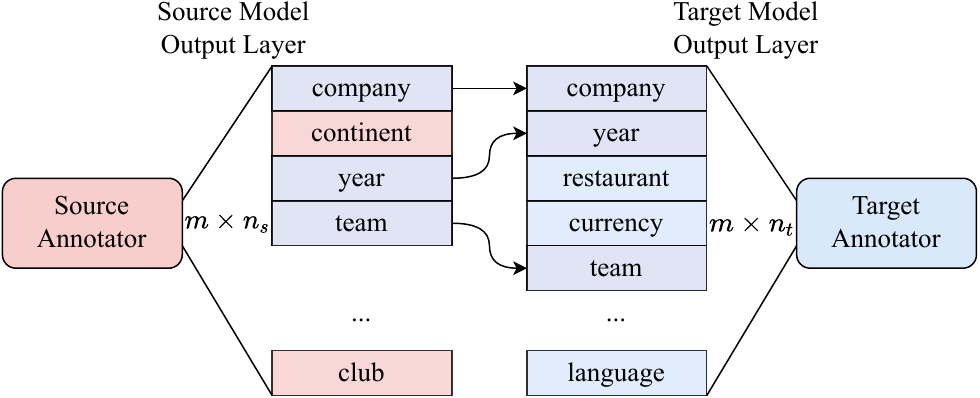}
   \vspace{-1.2em}
  \caption{Label set difference adjustment: output-layer columns for shared types are transplanted, others re-initialized.}
  \label{fig:label-diff}
    \vspace{-1.5em}
\end{figure}

This step addresses the label-set shift of \S\ref{sec:problem-cross}. Recall $L_s$ has shape $m \times n_s$ and is indexed by type. We transfer the encoder weights unchanged --- standard for reusing a fine-tuned representation~\cite{yosinski2014transferable} --- and rebuild the output layer type by type (Figure~\ref{fig:label-diff}): for each $s_{t,j} \in S_t$, if $s_{t,j}$ occurs in $S_s$ as $s_{s,i}$ we copy column $L_{s,i}$ into $L_{t,j}$, else initialize $L_{t,j}$ randomly. Types in $S_s \setminus S_t$ have no destination and are dropped, releasing capacity. Call the result $\Bar{M}_{t,0}$.

The transplant inherits as much source decision structure as the type sets permit, and \S\ref{sec:label-effect} shows it is decisive, not cosmetic --- though only a starting point, since a transplanted column encodes the \emph{source} boundary, which content shift or granularity mismatch may have invalidated. As $\Bar{M}_{t,0}$ has never seen $D_t$ and part of $L_t$ is random, we warm up on a small random sample $D_{f,0}$, avoiding a cold start where confidence estimates would be meaningless.

\subsubsection{Knowledge Difference Discovery} \label{sec:knowledge-diff}

We now localize the gap between $\Bar{M}_{t,l-1}$ and the annotator we want. At iteration $l$ we draw a random subset $D_l$ of not-yet-examined columns and, for each $C \in D_l$, take the pre-softmax output $v \in \mathbb{R}^{n_t}$ of $\Bar{M}_{t,l-1}$, whose confidence is the largest softmax score,
\[
\Phi(v) = \|\text{Softmax}(v)\|_{\infty} = \max_{i} \frac{e^{v_i}}{\sum_{k=1}^{n_t} e^{v_k}} \tag{1},
\]
and we place $C$ in the query set $A_l$ when $\Phi(v) < \delta$, leaving confident columns unqueried.

This filter is cost control, not the mechanism: querying all of $D_t$ would dominate the budget, and where the annotator is confident an LLM's generic knowledge is unlikely to overturn a domain-specific judgment. But low confidence alone is weak evidence, since neural classifiers are systematically overconfident off-distribution~\cite{guo2017calibration} --- precisely the regime here --- so $\Phi(v) < \delta$ over-selects, collecting genuine gaps together with merely borderline columns. Separating these is the LLM's job.

\stitle{Verification query.} For each $C_x \in A_l$ we prompt the LLM with a four-slot template: a task description instructing it to verify an annotation and answer only \textit{Yes}, \textit{No}, or \textit{I don't know}; the target type set $S_t$; the cells of $C_x$ concatenated into one string; and \emph{the label $\Bar{M}_{t,l-1}$ assigned to it}. The LLM returns only the decision $d_x$. Contrast prior work prompting ChatGPT to perform CTA~\cite{korini2023column}, where the LLM must select from a large type set: a multiple-choice question over $n_t$ options against our binary one. The supplementary material gives the verbatim template.

\stitle{Interpreting the decision.} If $d_x \in \{\text{No}, \text{I don't know}\}$ we add $C_x$ to the difficult set $\Tilde{A_l}$. The two answers mean different things, usefully so: ``No'' means the label is wrong by generic knowledge, so $C_x$ lies in $(T\setminus S)\cap G$ or in the part of $T \cap S$ needing realignment, while ``I don't know'' means the column exceeds generic knowledge, placing it in $(T\setminus S)\setminus G$ and reachable only via target supervision. Both merit labeling, hence the pooling. A ``Yes'' says the annotator was under-confident but right, so no budget need be spent. The LLM thus converts an uncalibrated confidence score into a decision about where knowledge falls short, at one cheap binary query per uncertain column.

Example~\ref{exam:cross} makes this concrete. On $C_{t,1}$ --- \textit{Euler}, \textit{Gauss} against the proposed \textit{Person} --- the answer is ``No'': the label is defensible in the ontology's terms yet wrong for $D_t$, which distinguishes \textit{Scientist}, and spotting that mismatch needs only generic knowledge ($T \cap S$ realignment). On $C_{t,3}$, university names, $M_s$ cannot propose \textit{University} at all and emits some in-vocabulary type at low confidence; rejecting it enters the column as $T \setminus S$ acquisition. Had it held a lake-specific code not inferable from values, the answer would be ``I don't know'' --- $(T \setminus S)\setminus G$, equally worth labeling since only target supervision resolves it. A correct but under-confident column draws ``Yes'' and costs nothing. The mechanism thus filters on \emph{where the annotator's competence ends}, not on the label itself.

\subsubsection{Why a Verifier and Not an Annotator}
\label{sec:why-verifier}

We claimed LLMs make poor CTA annotators, then built a component depending on one. The tension is only apparent, and resolving it is a design point rather than an implementation detail.

\stitle{Verification is strictly easier.} Annotation asks for the best of $n_t$ types ($78$ for VizNet, $275$ for Semtab2019); verification asks whether one proposed type is right. Selecting from a large ontology is hard even for human experts~\cite{wang2022sudowoodo}, and hard for a generic model for the same reason: it needs the lake's conventions, not just world knowledge. Verification needs only a compatibility judgment between values and a candidate type, closer to what pre-training supplies.

\stitle{The verifier cannot corrupt the output space.} In an LLM-as-annotator design the label \emph{is} the LLM's text, so nothing prevents a type outside $S_t$; measured OOD rates run $2.7\%$--$47.6\%$ (\S\ref{sec:no-tune}). In \sys the LLM emits no label: every prediction comes from the PLM output layer, defined over $S_t$ by construction (\S\ref{sec:background-plm}), so the OOD rate is zero however the LLM answers. The LLM influences only \emph{which columns get labeled}, where mistakes cost budget, not correctness. Failures thus degrade gracefully: a false ``Yes'' leaves a weak column unlabeled this round, recoverable as sampling continues, while a false ``No'' wastes labels on a column already handled but injects no error. Neither yields a wrong annotation --- the asymmetry making verification the safer role, and why a weaker verifier degrades performance gradually rather than breaking the pipeline (\S\ref{sec:add-robust}).

\subsection{Weak Sample Selection}
\label{sec:weak-sample}

$\Tilde{A_l}$ is too small and narrow to fine-tune on: it holds only columns from the sampled $D_l$, whereas the same weakness likely recurs across $D_t$ in columns never queried. We therefore \emph{propagate} each flagged column to its neighborhood, clustering all columns of $D_t$ with $K$-means~\cite{lloyd1982least, macqueen1967some} in the annotator's output-embedding space and marking a cluster $Q_k$ weak if it contains any column of $\Tilde{A_l}$. The union of weak clusters is the pool $D_w$, from which we draw a batch $D_{f,l}$ for annotation, decrementing the budget by $|D_{f,l}|$. Annotations come from the lake's labeling process --- human, heuristic, or knowledge-base-backed~\cite{langenecker2023steered, hulsebos2021making}; our experiments read ground-truth labels in its place.

Propagation rests on one load-bearing assumption: columns near a flagged column in the annotator's own embedding space are ones it treats similarly, hence likely difficult for the same reason. Clustering in the \emph{adapting model's} space rather than a fixed one is deliberate --- the geometry determining its errors is the geometry it currently has --- so we re-cluster each iteration. We set $K = n_t$: absent knowledge of the target label distribution, the number of types is the natural granularity, since ideally the lake decomposes into $n_t$ type-homogeneous groups. \S\ref{sec:parameter_sensitivity} shows performance peaks there and compares a Silhouette-selected $K$; \S\ref{sec:limitations} notes the assumption's limits.

\subsection{Gap-hopping Fine-tuning}
\label{sec:gap-hopping-finetuning}

Fine-tuning on $D_{f,l}$ alone would let the annotator drift toward the newest batch and forget earlier batches and transplanted knowledge --- catastrophic forgetting~\cite{robins1993catastrophic}. We adopt rehearsal~\cite{rebuffi2017icarl}: at iteration $l$ we fine-tune $\Bar{M}_{t,l-1}$ on the accumulated $\{D_{f,0}, \ldots, D_{f,l}\}$ for $N_f$ epochs, replaying every batch so far. Since $D_{f,0}$ is an unbiased random sample while later batches are weak-sample-biased, rehearsal also anchors the objective to the lake's overall distribution rather than only its hard regions. Types in $S_s \setminus T$, receiving no gradient, decay on their own.

\stitle{Iteration count and early stopping.} $P$ is an upper bound rather than a tuned operating point: early stopping terminates adaptation. Gains concentrate early, following from the LLM's role --- generic knowledge covers $(T\setminus S)\cap G$, and once absorbed the verifier has little left to offer, mirroring known LLM weakness on long-tail domains~\cite{sun2023head}. So when validation loss fails to improve for $N_e$ consecutive iterations we stop, keep the best checkpoint, and spend remaining budget on random unused columns; adaptation also halts if the budget is exhausted or the annotator is deemed sufficient. Several configurations stop well before $P$ (Table~\ref{tab:questions}), so $P$ mainly bounds worst-case query count. We fix $P=50$ throughout without per-transfer tuning.

\subsection{Complexity}
\label{sec:analysis}

Four steps add cost over fine-tuning the PLM alone. Realignment builds the source-to-target type mapping in $O(n_s n_t)$ and applies it in $O(n_t)$, so it is $O(n_s n_t)$ and runs once. Clustering costs $O(|D_t| n_t m I P)$~\cite{lloyd1982least, macqueen1967some} for embedding dimension $m$ ($768$ for BERT) and constant $K$-means iteration count $I$, the factor $P$ arising because we re-cluster each iteration. Rehearsal replays every batch collected so far, which with $F_1$ the fine-tuning cost of one column totals $O((P^2 N_D + N_t) F_1 N_f)$ for $N_D = \max\{|D_{f,0}|, |D_{f,l}|\}$, $l>0$; constraining $P N_D \le N_t$, which our settings satisfy, reduces this to $O(N_f P N_t F_1)$ --- exactly $P$ times ordinary fine-tuning's $O(N_f N_t F_1)$ --- and per-epoch validation is likewise $P$ times its single-pass cost. The LLM adds one binary query per low-confidence column. Every overhead is therefore one-off or a constant multiple of ordinary fine-tuning, and \S\ref{sec:time_money} confirms the absolute figures are small: under \textdollar$3$ and tens of minutes per transfer. The supplementary material gives pseudocode and the full derivation.

%% file: secs/sec_experiment.tex
\section{Experiments}
\label{sec:experiment}

Our evaluation addresses four claims, each checkable against specific evidence: \textbf{C1} \sys improves PLM-based annotators across lakes, especially under tight budgets and on long-tail types (\S\ref{sec:low-settings}--\S\ref{sec:gittables}); \textbf{C2} the improvement is attributable to LLM verification, not to confidence filtering or clustering alone (\S\ref{sec:ablation}); \textbf{C3} \sys matches fine-tuned table LLMs at a fraction of the adaptation cost and never emits out-of-domain labels (\S\ref{sec:no-tune}--\S\ref{sec:time_money}); \textbf{C4} gains do not depend on a particular verifier, backbone, or clustering hyperparameter (\S\ref{sec:add-robust}, \S\ref{sec:parameter_sensitivity}).

\subsection{Experimental Designs}

\subsubsection{Metrics, Datasets, Baselines, and Settings}
\stitle{Metrics.} We report Support-weighted F1 (\textbf{SW F1}), a support-weighted mean of per-type F1 reflecting overall accuracy and dominated by frequent types, and Macro-average F1 (\textbf{MA F1}), an unweighted mean in which rare types count equally, the more sensitive measure of long-tail behavior~\cite{zhang2019sato, sun2023reca}, plus \textbf{OOD rate} (\S\ref{sec:problem-cross}) for generative methods. Unless noted, numbers are F1 in $[0,1]$, higher better.

\stitle{Lakes and transfers.} Four lakes form three transfers spanning the regimes of \S\ref{sec:problem-cross}. PublicBI~\cite{vogelsgesang2018get} holds $33$ types over $160$ unusually wide tables ($64.6$ columns on average), VizNet~\cite{zhang2019sato} $78$ types over $32262$ narrow tables, Semtab2019~\cite{jimenez2020semtab} $275$ types over $3045$ tables, and GitTables~\cite{hulsebos2023gittables} comes from code repositories rather than the web; all are annotated against DBpedia but at different coverage and granularity. \textbf{P$\to$V} is pure type-set extension ($S_s \subset S_t$, nothing to forget); \textbf{V$\to$S} combines extension with forgetting, the type sets sharing only $13$ types ($16.7\%$ of VizNet, $4.7\%$ of Semtab2019); \textbf{P$\to$G} targets a corpus of altogether different provenance. We read ground-truth labels in place of the annotation step of Figure~\ref{fig:pipeline}, the contribution concerning which columns to annotate rather than how.

\stitle{Baselines.} PLM-based: Sherlock~\cite{hulsebos2019sherlock} fusing multi-granularity features, TABBIE~\cite{iida2021tabbie} with dual row/column transformers, DODUO~\cite{suhara2022annotating} encoding a table's columns jointly, Sudowoodo~\cite{wang2022sudowoodo} adding contrastive learning, and RECA~\cite{sun2023reca} retrieving related tables. The last three are the state of the art and the ones we adapt, spanning intra-table context, contrastive representations, and inter-table context, so improving all three is a stronger test than improving one. LLM-based: ChatGPT~\cite{korini2023column}, GPT-4o and GPT-5.1~\cite{achiam2023gpt}, and TableLlama~\cite{zhang2024tablellama}, which we also fine-tune at each budget.

\stitle{Settings.} Splits follow RECA~\cite{sun2023reca} where applicable, and every model trains to convergence on the source lake before adapting. We use Adam, cross-entropy loss, batch size $8$, sequence length $128$, $P=50$, $N_e=5$, $N_f=5$, and $\delta=0.9$, the last chosen by sweeping $[0,1]$ on $50$ held-out examples for highest filtering accuracy: a smaller $\delta$ misses weak samples, a larger one queries columns already handled. Warm-up sets are $50$ tables (P$\to$V) and $25$ (V$\to$S); per-iteration batches $D_{f,l}$ ($l>0$) are $25$ and $15$ columns. Hyperparameters are shared across transfers, not tuned per transfer. GPT-3.5-turbo-4k is \sys's default verifier, matching~\cite{korini2023column}. Experiments ran on four NVIDIA A800 80GB GPUs; the supplementary material gives full splits, per-baseline settings, and every remaining value.

\subsubsection{Our Approaches and Ablation Variants}
\label{sec:our-approaches}
We apply \sys to the three strongest backbones --- DODUO, Sudowoodo, RECA --- yielding \sys(D), \sys(S), \sys(R), and evaluate two variants, specified precisely since the distinction determines what the ablation establishes:

\stitle{$-$LLM verification.} The verifier is removed while \emph{every other component is retained}: confidence filtering still selects the low-confidence set (Eq.~1), $K$-means propagation still expands it, rehearsal is unchanged. The batch $D_{f,l}$ is drawn from clusters seeded by \emph{all} low-confidence columns rather than only the flagged subset $\Tilde{A_l}$. This is confidence-only active learning, the comparison isolating the verifier's marginal contribution --- hence we run it rather than a variant also dropping confidence filtering, which would conflate three components and support no claim about the LLM specifically.

\stitle{GPT-4o verification.} The verifier becomes GPT-4o, all else unchanged, testing sensitivity to verifier strength.

\subsubsection{Budgets}
\label{sec:plans}
Following~\cite{hedderich2020survey}, which notes the absence of a universal definition, we call budgets under $10\%$ of target training data \emph{low-resource} and above $10\%$ \emph{high-resource}, reporting both since they answer different questions: the low-resource regime is the deployment scenario the problem targets, where $N_t$ binds and column choice matters most, while the high-resource regime tests whether early gains \emph{survive} once labels are plentiful or merely front-load progress ordinary fine-tuning reaches anyway --- a method helping only in the first would be far less useful, as one cannot know in advance which regime a new lake falls into. Low-resource budgets equal the labels \sys consumes after 5, 10, 20, 30 iterations ($1.6\%$, $2.5\%$, $4.2\%$, $5.9\%$ for P$\to$V; $2.4\%$, $3.8\%$, $6.5\%$, $9.3\%$ for V$\to$S), all methods receiving the same budget; high-resource budgets are $25\%$, $50\%$, $100\%$.

\begin{table*}[hbtp]
\centering
  \caption{Low-resource results on three transfers. All values are F1 in $[0,1]$, higher better; column headers give the budget as a fraction of target training data and an absolute column count. Avg. Gain is the mean relative improvement over the corresponding backbone across the four budgets. Sherlock cannot run at the two smallest V$\to$S budgets, as its implementation requires every type to occur at least once in training.}
  \vspace{-1.1em}
  \label{tab:sota-viz}
    \scriptsize
    \setlength{\tabcolsep}{3.5pt}
  \begin{tabular}{lcccccccccc}
    \toprule
    \multicolumn{11}{c}{\textbf{(a) PublicBI $\to$ VizNet}}\\
    \multicolumn{1}{c}{}&\multicolumn{2}{c}{low1 1.6\% (239 col)}&\multicolumn{2}{c}{low2 2.5\% (364 col)}&\multicolumn{2}{c}{low3 4.2\% (614 col)}&\multicolumn{2}{c}{low4 5.9\% (864 col)}&\multicolumn{2}{c}{Avg. Gain}\\
     & SW F1 & MA F1 & SW F1 & MA F1& SW F1 & MA F1& SW F1 & MA F1 & SW F1 & MA F1\\
    \midrule
    Sherlock~\cite{hulsebos2019sherlock} & 0.344 & 0.130 & 0.470 & 0.238 & 0.558 & 0.303 & 0.591 & 0.345 & - & -\\
    TABBIE~\cite{iida2021tabbie} & 0.505 & 0.204 & 0.565 & 0.268 & 0.637 & 0.278 & 0.709 & 0.315 & - & -\\
    DODUO~\cite{suhara2022annotating} & 0.499 & 0.190 & 0.569 & 0.254 & 0.644 & 0.280 & 0.742 & 0.416 & - & -\\
    Sudowoodo~\cite{wang2022sudowoodo} & 0.561 & 0.213 & 0.601 & 0.277 & 0.705 & 0.374 & 0.724 & 0.427 & - & -\\
    RECA~\cite{sun2023reca} & 0.587 & 0.206 & 0.610 & 0.216 & 0.716 & 0.303 & 0.749 & 0.312 & - & -\\
    \midrule
    \sys(D) & 0.612 & 0.323 & 0.664 & 0.343 & 0.746 & 0.425 & 0.783 & 0.486 & 15.2\% $\uparrow$ & 43.4\% $\uparrow$\\
    \sys(S) & 0.609 & 0.317 & 0.679 & 0.384 & \textbf{0.776} & 0.446 & \textbf{0.814} & \textbf{0.558}& 11.0\% $\uparrow$ & 34.3\% $\uparrow$\\
    \sys(R) & \textbf{0.621} & \textbf{0.331} & \textbf{0.705} & \textbf{0.412} & 0.749 & \textbf{0.506} & 0.793 & 0.522 & 8.0\% $\uparrow$ & 71.4\% $\uparrow$\\
    \quad $-$LLM verification & 0.583 & 0.292 & 0.657 & 0.360 & 0.723 & 0.422 & 0.764 & 0.489 & - & - \\
    \quad GPT-4o verification & 0.614 & 0.329 & 0.669 & 0.345 & 0.750 & 0.427 & 0.786 & 0.489 & - & -\\
    \midrule
    \midrule
    \multicolumn{11}{c}{\textbf{(b) VizNet $\to$ Semtab2019}}\\
    \multicolumn{1}{c}{}&\multicolumn{2}{c}{low1 2.4\% (131 col)}&\multicolumn{2}{c}{low2 3.8\% (206 col)}&\multicolumn{2}{c}{low3 6.5\% (356 col)}&\multicolumn{2}{c}{low4 9.3\% (506 col)}&\multicolumn{2}{c}{Avg. Gain}\\
     & SW F1 & MA F1 & SW F1 & MA F1& SW F1 & MA F1& SW F1 & MA F1& SW F1 & MA F1\\
    \midrule
    Sherlock~\cite{hulsebos2019sherlock} & - & - & - & - & 0.225 & 0.096 & 0.313 & 0.154& - & -\\
    TABBIE~\cite{iida2021tabbie} & 0.322 & 0.097 & 0.375 & 0.127 & 0.494 & 0.210 & 0.580 & 0.287 & - & -\\
    DODUO~\cite{suhara2022annotating} & 0.302 & 0.110 & 0.393 & 0.164 & 0.521 & 0.245 & 0.594 & 0.303 & - & -\\
    Sudowoodo~\cite{wang2022sudowoodo} & 0.343 & 0.134 & 0.452 & 0.219 & 0.535 & 0.278 & 0.576 & 0.299& - & -\\
    RECA~\cite{sun2023reca} & 0.340 & 0.100 & 0.468 & 0.169 & 0.578 & 0.243 & 0.624 & 0.304& - & -\\
    \midrule
    \sys(D) & 0.365 & 0.144 & 0.475 & 0.215 & 0.573 & 0.296 & 0.620 & 0.370 & 14.0\% $\uparrow$ & 26.2\% $\uparrow$\\
    \sys(S) & 0.377 & \textbf{0.160} & \textbf{0.514} & \textbf{0.259} & 0.579 & 0.320 & 0.619 & 0.357 & 9.8\% $\uparrow$ & 18.0\% $\uparrow$\\
    \sys(R) & \textbf{0.424} & \textbf{0.160} & 0.503 & 0.248 & \textbf{0.633} & \textbf{0.394} & \textbf{0.672} & \textbf{0.429}& 12.3\% $\uparrow$ & 52.5\% $\uparrow$\\
    \quad $-$LLM verification & 0.350 & 0.131 & 0.423 & 0.176 & 0.524 & 0.264 & 0.604 & 0.322 & - & -\\
    \quad GPT-4o verification & 0.369 & 0.145 & 0.478 & 0.214 & 0.576 & 0.298 & 0.624 & 0.373 & - & -\\
    \midrule
    \midrule
    \multicolumn{11}{c}{\textbf{(c) PublicBI $\to$ GitTables} (budgets 10\%/25\%/50\%/100\% = 72/145/291/582 col; no Avg. Gain column)}\\
    \midrule
    Sherlock~\cite{hulsebos2019sherlock} & 0.272 & 0.212 & 0.420 & 0.335 & 0.506 & 0.388 & 0.523 & 0.400 & - & -\\
    TABBIE~\cite{iida2021tabbie} & 0.378 & 0.188 & 0.578 & 0.335 & 0.603 & 0.401 & 0.662 & 0.442 & - & -\\
    DODUO~\cite{suhara2022annotating} & 0.425 & 0.242 & 0.617 & 0.474 & 0.644 & 0.501 & 0.675 & 0.528 & - & -\\
    Sudowoodo~\cite{wang2022sudowoodo} & 0.451 & 0.287 & 0.577 & 0.430 & 0.661 & 0.522 & 0.688 & 0.531 & - & -\\
    RECA~\cite{sun2023reca} & 0.469 & 0.266 & 0.590 & 0.412 & 0.670 & 0.512 & 0.686 & 0.524 & - & -\\
    \midrule
    \sys(D) & 0.448 & 0.324 & \textbf{0.652} & \textbf{0.508} & 0.665 & \textbf{0.549} & \textbf{0.703} & \textbf{0.571} & - & -\\
    \sys(S) & 0.528 & 0.407 & 0.590 & 0.478 & \textbf{0.682} & 0.536 & 0.693 & 0.564 & - & -\\
    \sys(R) & \textbf{0.530} & \textbf{0.412} & 0.604 & 0.489 & 0.673 & 0.538 & 0.696 & 0.568 & - & -\\
    \bottomrule
  \end{tabular}
  \vspace{-1.2em}
\end{table*}

\begin{table*}[hbtp]
  \caption{High-resource results on the PublicBI$\to$VizNet and VizNet$\to$Semtab2019 transfers.}
  \vspace{-1em}
  \centering
  \label{tab:sota-full}
    \footnotesize
    \setlength{\tabcolsep}{3.5pt}
  \begin{tabular}{lcccccccccccc}
    \toprule
    \multicolumn{1}{c}{}&\multicolumn{6}{c}{PublicBI$\to$VizNet}&\multicolumn{6}{c}{VizNet$\to$Semtab2019}\\
    \multicolumn{1}{c}{}&\multicolumn{2}{c}{25\% (3745 col)}&\multicolumn{2}{c}{50\% (7490 col)}&\multicolumn{2}{c}{100\% (14980 col)}&\multicolumn{2}{c}{25\% (1363 col)}&\multicolumn{2}{c}{50\% (2725 col)}&\multicolumn{2}{c}{100\% (5450 col)}\\
     & SW F1 & MA F1 & SW F1 & MA F1& SW F1 & MA F1& SW F1 & MA F1& SW F1 & MA F1& SW F1 & MA F1\\
    \midrule
    Sherlock~\cite{hulsebos2019sherlock} & 0.714 & 0.465 & 0.759 & 0.482 & 0.791 & 0.592 & 0.494 & 0.292 & 0.567 & 0.341 & 0.637 & 0.424\\
    TABBIE~\cite{iida2021tabbie} & 0.825 & 0.470 & 0.849 & 0.501 & 0.862 & 0.562 & 0.692 & 0.446 & 0.712 & 0.455 & 0.765 & 0.568\\
    DODUO~\cite{suhara2022annotating} & 0.863 & 0.649 & 0.872 & 0.708 & 0.911 & 0.734 & 0.736 & 0.491 & 0.778 & 0.575 & 0.808 & 0.608\\
    Sudowoodo~\cite{wang2022sudowoodo} & 0.805 & 0.559 & 0.842 & 0.603 & 0.862 & 0.682 & 0.701 & 0.426 & 0.729 & 0.500 & 0.763 & 0.544\\
    RECA~\cite{sun2023reca} & 0.859 & 0.627 & 0.860 & 0.671 & 0.878 & 0.679 & 0.744 & 0.472 & 0.797 & 0.576 & 0.818 & 0.613\\
    \midrule
    \sys(D) & 0.874 & 0.661 & \textbf{0.902} & \textbf{0.749} & \textbf{0.927} & \textbf{0.791} & \textbf{0.753} & 0.538 & \textbf{0.798} & 0.612 & \textbf{0.819} & 0.622\\
    \sys(S) & \textbf{0.875} & \textbf{0.700} & 0.884 & 0.747 & 0.923 & 0.789 & 0.713 & 0.475 & 0.733 & 0.502 & 0.768 & 0.580\\
    \sys(R) & 0.860 & 0.658 & 0.871 & 0.700 & 0.880 & 0.705 & 0.748 & \textbf{0.545} & 0.797 & \textbf{0.618} & \textbf{0.819} & \textbf{0.643}\\
    \midrule
    \quad $-$LLM verification & 0.852 & 0.630 & 0.883 & 0.695 & 0.905 & 0.745 & 0.741 & 0.505 & 0.762 & 0.541 & 0.796 & 0.592 \\
    \quad GPT-4o verification & 0.876 & 0.664 & 0.904 & 0.751 & 0.929 & 0.792 & 0.757 & 0.546 & 0.799 & 0.614 & 0.820 & 0.621\\
    \bottomrule
  \end{tabular}
    \vspace{-1em}
\end{table*}

\subsection{Main Results (C1)}
\subsubsection{Low-resource Settings}
\label{sec:low-settings}
Table~\ref{tab:sota-viz} show that \sys improves all three backbones at every budget. On P$\to$V the mean relative gains over retraining the backbone directly are $15.2\%$, $11.0\%$, and $8.0\%$ in SW F1 and $43.4\%$, $34.3\%$, and $71.4\%$ in MA F1; on V$\to$S they are $14.0\%$, $9.8\%$, $12.3\%$ and $26.2\%$, $18.0\%$, $52.5\%$.

The gap between the metrics is more informative, and consistent with the mechanism rather than merely favorable: MA F1 gains exceed SW F1 gains two- to eight-fold, so most improvement comes from long-tail types. This follows from how columns are selected --- a backbone trained on few labels is weakest on rare types, so those are the low-confidence columns a verifier rejects and hence the ones entering the batch, whereas a uniformly spent budget would be dominated by frequent types already handled. The design predicts gains concentrate where support is thin, and they do. This is an argument from consistency, not a measurement of which types were selected (\S\ref{sec:limitations}).

\subsubsection{High-resource Settings}
\label{sec:high-settings}
Table~\ref{tab:sota-full} shows gains persisting at $25\%$--$100\%$ of training data, though \sys targets low budgets. Adaptation stops early here --- the iteration halts well before the budget is consumed, the remainder going to random columns --- so the improvement reflects a better early trajectory rather than continued LLM guidance. Gains are naturally smaller, and one configuration, \sys(S) on V$\to$S, falls below its backbone at the two largest budgets (\S\ref{sec:limitations}).

\subsubsection{A Third Transfer}
\label{sec:gittables}
P$\to$G tests whether this holds when the target is a different corpus entirely rather than a re-partition of web tables. Table~\ref{tab:sota-viz}(c) shows it does: all three variants improve over their backbones at every budget, MA F1 again gaining more than SW F1 ($+34\%$ for \sys(D) at $10\%$). Absolute scores are lower for every method, GitTables being harder, and backbone ordering changes --- \sys(R) and \sys(S) lead at the smallest budget, \sys(D) at the largest. The mechanism is thus not specific to one corpus pair, though \S\ref{sec:limitations} notes all three transfers remain unidirectional.

\subsection{Ablation: Is It the LLM? (C2)}
\label{sec:ablation}

Since $-$LLM verification retains confidence filtering, clustering, and rehearsal (\S\ref{sec:our-approaches}), its difference from full \sys is attributable to verification alone --- a comparison against confidence-only active learning, not random sampling. Across Tables~\ref{tab:sota-viz} and~\ref{tab:sota-full}, removing the verifier costs \sys(D) up to ${\sim}7\%$ relative in both metrics, more under tight budgets ($11\%$ MA F1 at V$\to$S low2) than at full data ($6\%$) --- what the design predicts, since when few labels can be bought which ones matters most, and confidence alone over-selects because a shifted classifier is overconfident (\S\ref{sec:knowledge-diff}).

The effect is not uniform, and we flag rather than omit the exception: at P$\to$V low2, $-$LLM verification reaches $0.360$ MA F1 against full \sys(D)'s $0.343$, the only reversal in $24$ metric-budget cells. Lacking variance over seeds we cannot separate a genuine effect from noise there, so we claim not that the verifier helps at every operating point but that it helps consistently in aggregate and most where the budget is tightest.

Replacing GPT-3.5 with GPT-4o changes little, improving SW F1 by under $0.005$ almost everywhere --- a useful negative result twice over: the mechanism needs no frontier verifier, and its benefit comes from asking a well-posed binary question rather than from raw capability (\S\ref{sec:add-robust} pushes this to an 8B open-weight model).

\subsection{Comparison with LLMs (C3)}

\subsubsection{Non-fine-tuned Performance}
\label{sec:no-tune}
Table~\ref{tab:chatgpt} shows prompted LLMs annotating below $0.5$ F1 on both lakes, under \sys at even its smallest budget. The OOD column is more consequential: ChatGPT and TableLlama place $26.4\%$ and $17.7\%$ of VizNet columns outside the type set, and $7.8\%$ and $47.6\%$ on Semtab2019, while GPT-5.1 reduces ($14.3\%$, $2.7\%$) but does not eliminate this. Fine-tuning suppresses without removing it: TableLlama still returns $1.7\%$--$2.9\%$ OOD labels at $100\%$ supervision. A pipeline receiving a label outside its schema cannot use it however plausible it reads, so this failure mode is qualitatively unlike misclassification. \sys's OOD rate is zero by construction (\S\ref{sec:why-verifier}).

\begin{table}[t!]
\centering
  \caption{LLM-based methods. SW F1/MA F1 are accuracy (higher better); OOD is the fraction of columns labeled outside $S_t$ (lower better). \sys has zero OOD by construction. TableLlama-\emph{x} is fine-tuned at budget \emph{x}.}
  \vspace{-1.2em}
  \label{tab:chatgpt}
  \footnotesize
  \setlength{\tabcolsep}{3pt}
  \begin{tabular}{lccc|ccc}
    \toprule
    & \multicolumn{3}{c|}{VizNet} & \multicolumn{3}{c}{Semtab2019} \\
    Methods & SW F1 & MA F1 & OOD & SW F1 & MA F1 & OOD\\
    \midrule
     ChatGPT~\cite{korini2023column} & 0.451 & 0.324 & 0.264 & 0.318 & 0.269 & 0.078 \\
    GPT-4o~\cite{achiam2023gpt} & 0.584 & 0.390 & 0.280 & 0.499 & 0.338 & 0.041 \\
    GPT-5.1~\cite{achiam2023gpt} & 0.686 & 0.453 & 0.143 & 0.572 & 0.401 & 0.027 \\
    TableLlama~\cite{zhang2024tablellama} & 0.146 & 0.086 & 0.177 & 0.113 & 0.078 & 0.476 \\
    \hdashline
     TableLlama-low1 & 0.620 & 0.349 & 0.027 & 0.552 & 0.386 & 0.021 \\
     TableLlama-low2 & 0.694 & 0.395 & 0.029 & 0.541 & 0.412 & 0.018 \\
     TableLlama-low3 & 0.749 & 0.482 & 0.025 & 0.590 & 0.440 & 0.014 \\
    TableLlama-low4 & 0.793 & 0.507 & 0.027 & 0.645 & 0.467 & 0.021 \\
    TableLlama-25\% & 0.882 & 0.591 & 0.019 & 0.755 & 0.570 & 0.020 \\
    TableLlama-50\% & 0.903 & 0.608 & 0.018 & 0.802 & 0.647 & 0.009 \\
     TableLlama-100\% & 0.925 & 0.642 & 0.017 & 0.846 & 0.702 & 0.009\\
  \bottomrule
   \end{tabular}
   \vspace{-1.6em}
\end{table}

\subsubsection{Fine-tuned Performance}
\label{sec:tune}
We fine-tuned TableLlama~\cite{zhang2024tablellama} at each budget. Comparing Table~\ref{tab:chatgpt} against Tables~\ref{tab:sota-viz} and~\ref{tab:sota-full}, its F1 is broadly comparable to \sys's at matched budgets, so we claim no superior annotation quality. The difference is cost: fine-tuning throughput is $1.43$ columns/s against $39.87$, $85.03$, $188.15$ for \sys(R), (S), (D) --- $27.9\times$, $59.5\times$, $131.6\times$ faster --- plus far smaller memory needs, and \sys differs in kind on the output space, TableLlama emitting out-of-schema labels at every budget. To be explicit: the cross-table F1 comparison is a rough equivalence claim at matched budgets, not a ranking; the strict claims are on adaptation speed and OOD rate.

\subsection{Comprehensive Experiments (C3, C4)}
\label{sec:analyses}

\subsubsection{Time and Monetary Overhead} \label{sec:time_money}
\S\ref{sec:analysis} bounds rehearsal overhead at a constant multiple of ordinary fine-tuning; Table~\ref{tab:questions} quantifies the rest. End to end, P$\to$V costs \sys(D), (S), (R) $1704.3$s/\textdollar$0.64$, $1776.3$s/\textdollar$0.66$, $1960.0$s/\textdollar$0.73$ and V$\to$S $1511.6$s/\textdollar$2.26$, $1726.4$s/\textdollar$2.58$, $1189.4$s/\textdollar$1.78$. Semtab2019's per-query cost roughly triples VizNet's, enumerating $275$ types costing more prompt tokens than $78$, so overhead scales with type-set size. Clustering adds under $70$s total, and all overheads stay under \textdollar$3$ and tens of minutes per transfer.

\begin{table}[t!]
  \centering
  \caption{Left: LLM queries and realized iterations (early stopping halts several runs before $P{=}50$). Right: cost per query-response pair.}
  \vspace{-1.2em}
  \label{tab:questions}
  \footnotesize
  \setlength{\tabcolsep}{2.5pt}
  \begin{tabular}{llcc}
    \toprule
    Transfer & Method & \# queries & \# iter.\\
    \midrule
    \multirow{3}*{P$\to$V} & \sys(D) & 1348 & 38\\
    ~ & \sys(S) & 1405 & 45\\
    ~ & \sys(R) & 1550 & 50\\
    \midrule
    \multirow{3}*{V$\to$S} & \sys(D) & 1506 & 50\\
    ~ & \sys(S) & 1720 & 50\\
    ~ & \sys(R) & 1185 & 38\\
    \bottomrule
  \end{tabular}
  \begin{tabular}{lcc}
    \toprule
    Dataset & \textdollar/query & s/query\\
    \midrule
    VizNet & 0.00047 & 1.2643 \\
    Semtab2019 & 0.00150 & 1.0037\\
    \bottomrule
  \end{tabular}
  \vspace{-1.6em}
\end{table}

\subsubsection{Reliability of LLM Verification}
\label{sec:reliability-veri}
\S\ref{sec:why-verifier} argued verification is easier than annotation. On identical columns, ChatGPT prompted to annotate is correct on $0.424$ (VizNet) and $0.339$ (Semtab2019) of columns; prompted to verify its proposed label, on $0.807$ and $0.870$, with rare abstentions ($0.009$, $0.007$).

The comparison's limits bear on how much it supports. Chance levels differ ($1/n_t$ versus $1/2$), so part of the gap is mechanical, and the numbers show verification is easier without showing it \emph{well posed}. Accuracy is also base-rate sensitive: if most queried labels were already correct, an indiscriminate ``Yes'' verifier would score well while carrying no information. Since the pipeline consumes only the ``No''/``I don't know'' decisions, the informative quantities are that base rate and the precision and recall of ``No'' decisions, which we do not report. We therefore limit the claim: verification is tractable for a generic LLM, while the ablation of \S\ref{sec:ablation} --- not these numbers --- evidences that the decisions carry signal (\S\ref{sec:limitations}).

\subsubsection{Effect of Label Set Difference Adjustment}
\label{sec:label-effect}
Measuring the annotator right after realignment and warm-up, before any adaptation iteration, isolates the transplant of \S\ref{sec:label-diff}. With a randomly initialized output layer it is essentially inoperative, SW F1 at or below $0.028$ across all three backbones on both transfers; with the transplant, \sys(D) and \sys(S) reach $0.241$ and $0.231$ on P$\to$V and roughly $0.09$ on V$\to$S. Absolute values are low, almost no target supervision having been consumed, so these compare initializations rather than end performance. \sys(R) benefits least on P$\to$V ($0.010$), consistent with RECA's reliance on inter-table context a transplanted output layer does not supply.

\subsubsection{Parameter Sensitivity}
\label{sec:parameter_sensitivity}
Varying $K \in \{1, 0.5n_t, n_t, 1.5n_t, 2n_t\}$ over the low-resource budgets with GPT-4o as verifier, performance peaks at $K=n_t$ on both transfers and in both metrics, matching the rationale of \S\ref{sec:weak-sample}: with roughly $n_t$ type-homogeneous groups, that granularity propagates a flagged column to columns sharing its difficulty without absorbing unrelated ones. Degradation away from the peak is gradual, so the choice is not brittle. Selecting $K$ by Silhouette score~\cite{rousseeuw1987silhouettes} instead gives $K=255$ against $n_t=275$ on Semtab2019, and the two lie within $0.02$ F1 at every V$\to$S budget while differing inconsistently in direction --- unsurprising, as the score itself moves only from $0.984$ to $0.987$, leaving the partitions nearly identical. We therefore treat $K=n_t$ as a default that costs no extra computation, and suggest computing the score only when the target type distribution is severely imbalanced.

\subsubsection{Robustness}
\label{sec:add-robust}
We replace \emph{both} ends of the pipeline on P$\to$V, using Llama-3.1-8B-Instruct as verifier and RoBERTa as encoder. This reaches $0.718$/$0.754$/$0.807$/$0.827$ SW F1 across the four low-resource budgets and $0.858$/$0.882$/$0.919$ at $25\%$/$50\%$/$100\%$, with MA F1 $0.370$/$0.453$/$0.516$/$0.514$ and $0.610$/$0.694$/$0.789$: against \sys(D) in Tables~\ref{tab:sota-viz} and~\ref{tab:sota-full}, gains over the backbone persist throughout. The 8B result matters beyond robustness: the verifier need not be a hosted frontier model, so \sys can run wholly on premises where sending column values to a third party is unacceptable.

\subsubsection{What Governs Adaptation Difficulty}
\label{sec:add-discuss}
Two factors drive how much target data a transfer needs. The backbone: to pass $0.6$ SW F1 on V$\to$S, \sys(D) and \sys(S) need $506$ columns against \sys(R)'s $356$, which we attribute to RECA's inter-table context supplying signal otherwise bought with labels. And the transfer: adaptation is faster on P$\to$V than V$\to$S, consistent with the former being pure type-set extension and the latter also requiring forgetting (\S\ref{sec:problem-cross}), and slower still on P$\to$G. Both are post-hoc explanations rather than controlled comparisons, since backbone and transfer cannot be varied independently of available datasets.

%% file: secs/sec_limitations.tex
\section{Limitations and Threats to Validity}
\label{sec:limitations}

We separate what our experiments establish from what they do not, naming the evidence that would close each gap.

\stitle{Residual error, characterized.} Reporting only gains would misrepresent the problem's state, so we bound what remains. \emph{The long tail is improved, not solved}: MA F1 is both where \sys gains most and where absolute performance stays weakest, reaching $0.32$ on P$\to$V but only $0.14$ on V$\to$S at the tightest budgets, against SW F1 of $0.61$ and $0.37$, and topping out at $0.79$ and $0.64$ even under full supervision --- so the metric gap \S\ref{sec:low-settings} reads as tail-directed selection equally measures how much tail error survives. \emph{Harder targets narrow the margin}: every method scores lower on P$\to$G than P$\to$V, at the smallest V$\to$S budget \sys(D) reaches $0.365$ SW F1 --- a starting point for human curation, not an autonomous annotator --- and cost grows with type-set size (\S\ref{sec:time_money}). \emph{And one configuration regresses}: \sys(S) on V$\to$S falls below its backbone at $50\%$ and $100\%$, while $-$LLM verification beats full \sys(D) in one of $24$ low-resource cells. We keep both in the tables rather than reporting aggregates alone: \sys reliably improves the low-budget regime it targets, while guaranteeing nothing at an individual high-budget operating point.

\stitle{Weak-sample selection is validated end to end, not mechanistically.} The evidence that flagged columns correspond to genuine gaps (\S\ref{sec:weak-sample}) is indirect: removing the verifier costs accuracy (\S\ref{sec:ablation}), gains concentrate in MA F1 as a rare-type-biased selector predicts (\S\ref{sec:low-settings}), and the decision semantics map onto the three regions case by case (\S\ref{sec:knowledge-diff}) --- but none inspects the columns actually selected. A direct test would report the composition of $D_{f,l}$ against ground truth: what fraction lies in $S_t \setminus S_s$, in the long tail, or in realignment-requiring shared types, relative to the lake's marginal distribution. Two caveats compound the omission: confidence and embeddings both come from a model still adapting, so the selector is judged by the representation it should repair, and we did not test a frozen-space clustering variant. Our claim is thus that selection improves adaptation, not that it provably targets the three regions of \S\ref{sec:problem-cross}.

\stitle{Forgetting is assessed by aggregate accuracy, not per-region.} Rehearsal (\S\ref{sec:gap-hopping-finetuning}) should retain $T \cap S$ while releasing $S \setminus T$ and acquiring $T \setminus S$; aggregate target F1 confirms the composite outcome without separating the three effects. The right evaluation would decompose accuracy by knowledge region --- per-type F1 on shared types across iterations to detect erosion, and whether residual $S \setminus T$ mass still attracts probability --- and probe source-lake retention, since a practitioner serving both lakes cares whether adaptation destroys $M_s$. We report neither, though the residual error bounded above is what any such decomposition must explain. Note also that rehearsal replays only \emph{target} batches, so nothing explicitly protects shared knowledge beyond the transplanted initialization, and on P$\to$V forgetting cannot arise at all, leaving V$\to$S and P$\to$G to carry this evidence.

\stitle{Scope: three unidirectional transfers, one source.} Our transfers span pure extension, extension with forgetting, and a change of corpus, but remain a small, structurally narrow sample. We run no reverse direction, so cannot say whether adaptation is symmetric; the reverse of P$\to$V, namely $S_t \subset S_s$, is a pure-forgetting regime our design has never faced. Nor do we test disjoint type sets, where the transplant of \S\ref{sec:label-diff} would have nothing to copy and \sys should degrade toward ordinary low-resource fine-tuning, or hierarchical and multi-label type sets. We also assume the source annotator is given, whereas a target lake may admit several candidates; choosing or combining them likely matters as much as the adaptation procedure, since the starting backbone substantially changes the required budget (\S\ref{sec:add-discuss}).

\stitle{Verification accuracy without base rates, and no variance.} The $0.807$/$0.870$ accuracies of \S\ref{sec:reliability-veri} are uncorrected for the base rate of correct labels in the query set, and the formulations differ in chance level; an indiscriminate ``Yes'' verifier would score highly while contributing nothing. Missing are that base rate and the precision and recall of ``No'' decisions, since only those drive labeling, so we rest the claim that verification carries usable signal on the end-to-end ablation instead. Separately, all results are single runs, so we cannot attach confidence intervals or separate small differences from seed noise --- which matters for the isolated reversal at P$\to$V low2 and the GPT-3.5/GPT-4o near-ties, less so for headline comparisons whose gaps are large and consistent in sign across three transfers, three backbones, and seven budgets.

\stitle{Component novelty versus system novelty.} The individual mechanisms are established --- output-layer transplantation~\cite{yosinski2014transferable}, confidence-based uncertainty sampling~\cite{settles2009active}, $K$-means expansion, rehearsal~\cite{robins1993catastrophic, rebuffi2017icarl} --- and we do not claim otherwise. We claim the knowledge decomposition determining \emph{which} mechanism addresses which part of the gap, and the verifier role making generic knowledge usable without surrendering the closed output space: a property no combination of those components alone provides, and why \sys attains zero OOD rate where LLM annotators reach $47.6\%$. Whether that suffices as novelty we leave to the reader, having made its basis explicit.

%% file: secs/sec_related.tex
\section{Related Work}
\label{sec:related}

\stitle{PLM-based CTA.} PLMs such as BERT~\cite{devlin2019bert} enabled expressive table representations. TaBERT~\cite{yin2020tabert} and TABBIE~\cite{iida2021tabbie} first used them to encode table content; DODUO~\cite{suhara2022annotating} encodes all columns of a table jointly for intra-table semantics; Sudowoodo~\cite{wang2022sudowoodo} adds contrastive pre-training for label scarcity; RECA~\cite{sun2023reca} retrieves structurally related tables for inter-table context.

These lines improve the \emph{representation} a fixed lake induces but do not address the transition between lakes: each assumes a fixed type set determining the output layer's shape, so none can express a target type absent from its source lake, and each treats a new lake as fresh fine-tuning where every label is bought anew. Better representations also do not tell a practitioner \emph{which} columns to annotate first --- the cost dominating adaptation. \sys is therefore complementary, operating on the output layer and training-data selection while leaving the encoder untouched, which is why one procedure lifts three architecturally distinct backbones (\S\ref{sec:low-settings}).

\stitle{LLM-based CTA.} LLMs~\cite{achiam2023gpt, touvron2023llama} allow CTA as question answering. \citet{korini2023column} design prompts asking ChatGPT to pick a column's type, obtaining zero-shot cross-lake generalization with no training, but our measurements show the annotations are weak on long-tail types and frequently fall outside the requested type set (\S\ref{sec:no-tune}). Later work fine-tunes LLMs toward PLM-level accuracy~\cite{feuer2023archetype, li2024table, zhang2024tablellama} at a compute and memory cost forfeiting the cheap portability that motivated an LLM: our comparison puts adaptation $27$--$131\times$ slower than \sys.

The open design question is not whether to use an LLM but \emph{in what role}. Prior work places it in the output path, which is what admits out-of-schema answers; \sys places it in the control path, deciding which columns deserve annotation while predictions come from a classifier over $S_t$. Generic knowledge thus judges plausibility without committing to a type in a lake's specific ontology, yielding a combination neither family offers: portability, PLM-level accuracy, and a structurally closed output space (Table~\ref{tab:comparison}).

\stitle{Cross-lake adaptation and data programming.} SigmaTyper~\cite{hulsebos2021making} articulated the vision of reusing annotators across lakes; STEER~\cite{langenecker2023steered} pursues it by generating target labels from expert-written labeling functions and retraining. STEER attacks the \emph{supply} of labels, \sys their \emph{allocation} under a budget, and the two compose: STEER's labeling functions can serve as the annotation process our selection step invokes.

\stitle{Transfer and active learning.} Our mechanisms draw on parameter transfer between related tasks~\cite{yosinski2014transferable, gururangan2020don, zhuang2020comprehensive}, uncertainty and diversity sampling~\cite{settles2009active}, and rehearsal against forgetting~\cite{robins1993catastrophic, rebuffi2017icarl}. Transfer over tabular data is harder than over text owing to heterogeneity~\cite{jain2021deep, levin2022transfer}, and table applications have largely targeted value prediction~\cite{wang2022transtab, levin2022transfer} rather than schema-level understanding. Our contribution is not a new sampling or rehearsal algorithm but the decomposition assigning each a specific part of the cross-lake gap, plus the observation that an LLM supplies the signal for the part generic knowledge covers.

%% file: secs/sec_conclusion.tex
\section{Conclusion}
\label{sec:conclusion}

We studied how to move a column type annotator between data lakes under a fixed annotation budget, arguing the task is knowledge management: a source annotator holds knowledge to discard, to realign and reuse, and to acquire, and intersecting these regions with a generic model's knowledge determines what an LLM can supply. \sys casts the LLM as a \emph{verifier} rather than an annotator, which borrows generic knowledge for deciding where to spend labels while keeping every prediction inside the target type set, so unlike prompted LLMs it emits no out-of-schema labels. Across three transfers and three backbones it nears full-data quality with under $6\%$ of target labels, adapting $27$--$131\times$ faster than a fine-tuned table LLM. \S\ref{sec:limitations} states where it falls short; reverse and disjoint-type-set transfers, hierarchical type sets, and source selection are the natural next steps.

%% file: supplementary_arxiv.tex

\noindent This appendix provides material that supplements the main paper: the verification prompt template (Appendix~\ref{sec:supp-verification}), the full algorithm and its complexity derivation (Appendix~\ref{sec:supp-algorithm}), complete experimental preliminaries including dataset statistics, baseline descriptions, and every hyperparameter (Appendix~\ref{sec:supp-preli}), and the detailed tables and figures behind the analyses reported numerically in the main text (Appendix~\ref{sec:supp-results}).

\section{LLM Verification Template}
\label{sec:supp-verification}

\begin{figure}[hbtp]
    \centering
    \includegraphics[width=0.95\linewidth]{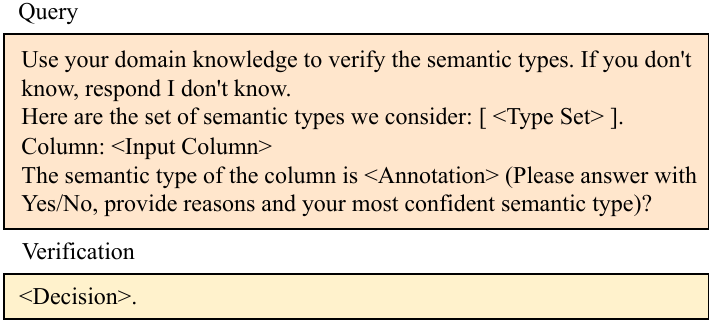}
    \vspace{-1em}
    \caption{The LLM verification query template.}
    \label{fig:template}
\end{figure}

Figure~\ref{fig:template} gives the prompt template used in the knowledge difference discovery step (\S\ref{sec:knowledge-diff}). It has four slots. The task description instructs the model to verify an annotation and to answer only \textit{Yes}, \textit{No}, or \textit{I don't know}. $<$Type Set$>$ enumerates the target type set $S_t$. $<$Input Column$>$ receives the cell values of the queried column $C_x$, concatenated into a single string. $<$Annotation$>$ receives the label that the current annotator $\Bar{M}_{t,l-1}$ assigned to that column. The model returns only $<$Decision$>$.

The contrast with prompting an LLM to \emph{annotate}~\cite{korini2023column} is the point of the design. That template asks the model to select the most appropriate type from $S_t$, a multiple-choice question over $78$ or $275$ options; ours asks whether one proposed type is correct, a binary question. Selecting from a large ontology is difficult even for human experts~\cite{wang2022sudowoodo}, and difficult for a generic model for the same reason: it requires the target lake's annotation conventions rather than world knowledge alone. \S\ref{sec:why-verifier} develops this argument and \S\ref{sec:reliability-veri} measures the resulting accuracy difference.

\section{Algorithm and Complexity}
\label{sec:supp-algorithm}

Algorithm~\ref{algo:Lakehopper} gives the pseudocode for \sys. Lines 1--5 perform the one-off label set difference adjustment; lines 8--29 are the adaptation loop, in which lines 11--14 filter by confidence, lines 15--18 query the verifier, lines 19--22 propagate flagged columns into a labeling batch, and lines 23--28 fine-tune with rehearsal and test the early-stopping condition. Line 30 spends any residual budget on randomly sampled unused columns.

\begin{algorithm}[t!]
\caption{\sys}
\label{algo:Lakehopper}
{\small
\begin{algorithmic}[1]{\color{black}
\REQUIRE ~~\\
Maximum number of iterations $P$\\
Target data lake $D_t$\\
Number of fine-tuning epochs in each iteration $N_f$\\
Early-stopping threshold $N_e$\\
Annotation budget $N_t$\\
Confidence threshold $\delta$\\
Source annotator $M_s$\\
Source label set $S_s$\\

\ENSURE ~~\\
The adapted target annotator $M_t$
\STATE Copy the weights of $M_s$ except the output layer $L_s$ to the intermediate target annotator $\Bar{M}_{t,0}$.
\FOR{$j=1,2,\ldots,n_t$}
\STATE \textbf{If} $s_{t,j} \in S_s$ and $s_{t,j}=s_{s,i}$ \textbf{then}, assign $L_{s,i}$ to $L_{t,j}$.
\STATE \textbf{Otherwise}, randomly initialize $L_{t,j}$.
\ENDFOR
\STATE Randomly sample a subset $D_{f,0}$ from $D_t$ to warm up $\Bar{M}_{t,0}$; update the budget $N_t = N_t-|D_{f,0}|$.
\STATE Initialize the best validation loss $\Bar{v}=\infty$ and the number of non-improving iterations $\alpha=0$.
\FOR{$l=1,2,\ldots,P$}
\STATE Initialize $A_l=\emptyset$ and $\Tilde{A_l}=\emptyset$.
\STATE Randomly sample a subset $D_l$ of columns not yet sampled from $D_t$.
\FOR{each column $C$ in $D_l$}
\STATE Obtain the output embedding $v$ with $\Bar{M}_{t,l-1}(C)$.
\STATE \textbf{If} $\Phi(v) < \delta$ \textbf{then}, append $C$ to $A_l$.
\ENDFOR
\FOR{each column $C$ in $A_l$}
\STATE Query the LLM with $C$ and obtain the decision $d_x$.
\STATE \textbf{If} $d_x=$ `No' or `I don't know' \textbf{then}, add $C$ to $\Tilde{A_l}$.
\ENDFOR
\STATE Initialize the weak sample pool $D_w=\emptyset$.
\STATE Perform $K$-means clustering on $D_t$ to obtain clusters $Q_k$.
\STATE Include $Q_k$ in $D_w$ if it contains any column in $\Tilde{A_l}$.
\STATE Randomly sample a subset $D_{f,l}$ from $D_w$ for annotation; update the budget $N_t=N_t-|D_{f,l}|$.
\STATE Fine-tune $\Bar{M}_{t,l-1}$ with $\{D_{f,0}, D_{f,1}, \ldots, D_{f,l}\}$ for $N_f$ epochs to obtain $\Bar{M}_{t, l}$.
\STATE Compute the validation loss $\Bar{v}_l$ of $\Bar{M}_{t,l}$.
\STATE \textbf{If} $\Bar{v}_l < \Bar{v}$ \textbf{then}, assign $\Bar{v} = \Bar{v}_l$, $\alpha=0$, $M_t=\Bar{M}_{t,l}$.
\STATE \textbf{Otherwise}, assign $\alpha = \alpha + 1$.
\IF{$\alpha \ge N_e$}
\STATE break
\ENDIF
\ENDFOR
\STATE Fine-tune $M_t$ on $N_t$ randomly sampled unused columns from $D_t$.
\RETURN $M_t$.}
\end{algorithmic}
} 
\end{algorithm}

\stitle{Complexity.} Relative to fine-tuning the PLM alone, the overheads are the label set difference adjustment (lines 1--5), the LLM queries (lines 15--18), the $K$-means clustering (lines 19--21), and the additional passes induced by rehearsal (lines 6, 23). Query time and monetary cost are measured in \S\ref{sec:time_money}; we account for the rest here.

For the label set difference adjustment we build a source-to-target type mapping, at a cost of $O(n_s n_t)$. Given the mapping, rebuilding the output layer costs $O(n_t)$, since we consult the mapping once per column of $L_t$. The step is therefore $O(n_s n_t)$ overall and runs once.

The $K$-means clustering step takes $O(|D_t| n_t m I P)$~\cite{lloyd1982least, macqueen1967some}, where $m$ is the output embedding dimension ($768$ for BERT), $I$ is the number of $K$-means iterations, which is constant, and the factor $P$ arises because we re-cluster at every adaptation iteration.

For rehearsal, let $F_1$ denote the fine-tuning cost of a single column. Replaying every batch collected so far gives
\begin{align*}
&O\Big(\big(P|D_{f,0}|+\tfrac{P(P-1)}{2}|D_{f,l}| + N_t-(P-1)|D_{f,l}|-|D_{f,0}|\big) F_1 N_f\Big)\\
&= O\big((P^2 N_D + N_t) F_1 N_f\big),
\end{align*}
where $N_D = \max\{|D_{f,0}|, |D_{f,l}|\}$ for $l>0$. Constraining $P$ and $N_D$ so that $P N_D \le N_t$, which our settings satisfy, yields an overall $O(N_f P N_t F_1)$: exactly $P$ times the annotator's ordinary fine-tuning cost of $O(N_f N_t F_1)$. A further overhead comes from validation at each epoch. Writing $F_2$ for the validation cost of a single column and $N_v$ for the size of the validation set, this is $O(P N_f F_2 N_v)$, again $P$ times the original $O(N_f F_2 N_v)$.

Finally, the LLM adds one binary query per low-confidence column. Every overhead is therefore either one-off or a constant multiple of ordinary fine-tuning, and \S\ref{sec:time_money} confirms that the absolute magnitudes are small: under \textdollar$3$ and tens of minutes per transfer.

\section{Experimental Preliminaries}
\label{sec:supp-preli}

\subsection{Metrics}
\label{sec:supp-metrics}

We evaluate with Support-weighted F1 (SW F1) and Macro-average F1 (MA F1)~\cite{zhang2019sato, sun2023reca}. SW F1 is a weighted average of per-type F1 scores with weights given by each type's support, so it tracks overall accuracy and is dominated by frequent types. MA F1 is the unweighted mean of per-type F1 scores, so rare types contribute equally and it is the more sensitive measure of long-tail behavior. Both lie in $[0,1]$ and higher is better. For generative methods we additionally report the out-of-domain (OOD) rate, defined in \S\ref{sec:problem-cross} as the fraction of predictions falling outside $S_t$.

\subsection{Datasets}
\label{sec:supp-datasets}

We use four real-world datasets as data lakes: PublicBI~\cite{vogelsgesang2018get}, VizNet~\cite{zhang2019sato}, Semtab2019~\cite{jimenez2020semtab}, and GitTables~\cite{hulsebos2023gittables}, following prior work~\cite{sun2023reca, langenecker2023steered, zhang2019sato}. From VizNet we take the multi-table-only subset of the WebTables corpus, following~\cite{zhang2019sato, sun2023reca}; from Semtab2019 we take the same subset as RECA~\cite{sun2023reca}.

Table~\ref{tab:dataset} reports statistics for the first three. PublicBI contains the fewest tables and its tables are much wider than those of the others. VizNet is the largest and its tables tend to be narrow. All are annotated with the DBpedia ontology, but they cover different aspects of it at different granularity levels. We consider this spread representative of real deployments: different lakes hold tables of different sizes and content, annotated against different portions of an ontology.

\begin{table}[h]
  \caption{Statistics of the PublicBI, VizNet, and Semtab2019 datasets.}
  \centering
  \small
  \vspace{-1em}
  \begin{tabular}{c|ccc}
    \hline
    Datasets & PublicBI & VizNet & Semtab2019\\
    \hline
    \# types & 33 & 78 & 275\\
    \# tables & 160 & 32262 & 3045\\
    \# annotated columns & 1424 & 74141 & 7603\\
    Avg. \# columns & 64.6 & 2.3 & 4.5 \\
    \hline
  \end{tabular}
  \label{tab:dataset}
\end{table}

\begin{figure}[h]
  \centering
  \includegraphics[width=0.86\linewidth]{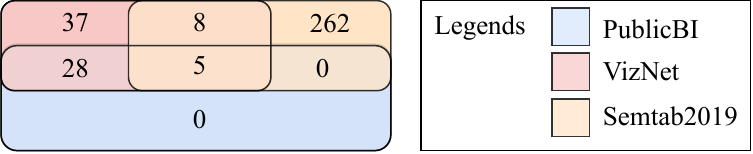}
  \vspace{-1em}
  \caption{Type set overlap among the three datasets.}
  \label{fig:venn-dataset}
  \vspace{-0.5em}
\end{figure}

These lakes yield the three transfers evaluated in the paper, and Figure~\ref{fig:venn-dataset} shows the type set overlaps that make them differ in kind. PublicBI's type set is a strict subset of VizNet's, so \textbf{PublicBI$\to$VizNet} (P$\to$V) is pure type-set extension: every source type survives and the annotator need only acquire the additional target types. VizNet and Semtab2019 overlap in only $13$ types, $16.7\%$ of VizNet's and $4.7\%$ of Semtab2019's, so \textbf{VizNet$\to$Semtab2019} (V$\to$S) additionally requires forgetting source-only types while learning target-only ones. \textbf{PublicBI$\to$GitTables} (P$\to$G) targets a corpus of different provenance altogether rather than a re-partition of web tables. In place of the annotation step of the pipeline we read the ground-truth labels supplied by the datasets, since the paper's contribution concerns which columns to annotate rather than how annotation is performed.

The licenses of PublicBI, VizNet, Semtab2019, and GitTables are the MIT License, CC BY 4.0, Apache License 2.0, and CC BY 4.0 respectively. We followed them throughout.

\subsection{Baselines}
\label{sec:supp-baselines}

\textbf{Sherlock}~\cite{hulsebos2019sherlock} fuses multi-granularity features (characters, words, paragraphs, and global context). \textbf{TABBIE}~\cite{iida2021tabbie} encodes columns and rows with dual transformers. \textbf{DODUO}~\cite{suhara2022annotating} jointly encodes all columns of a table with a transformer. \textbf{Sudowoodo}~\cite{wang2022sudowoodo} adds contrastive learning for low-resource settings. \textbf{RECA}~\cite{sun2023reca} discovers structurally related tables to supply inter-table context.

DODUO, Sudowoodo, and RECA are the state of the art and the three backbones we adapt. They also span the relevant design space: DODUO exploits intra-table context, Sudowoodo contrastive representations, and RECA inter-table context. DODUO and RECA are reported label-efficient~\cite{suhara2022annotating, sun2023reca}, and Sudowoodo targets label scarcity by construction. Improving all three is therefore a stronger test than improving any one of them.

LLM-based baselines are ChatGPT prompted for CTA~\cite{korini2023column}, GPT-4o and GPT-5.1~\cite{achiam2023gpt}, and TableLlama~\cite{zhang2024tablellama}, which we additionally fine-tune at each budget.

\subsection{Settings}
\label{sec:supp-settings}

\stitle{Splits.} For PublicBI we randomly select $10\%$ of the labeled data as the test set and use the rest for training. For VizNet we select three of five folds from the WebTables corpus, using one each for training, validation, and testing. For Semtab2019 we follow RECA~\cite{sun2023reca}: $10\%$ of annotated columns form the test set and the remainder is split $1{:}4$ into validation and training. Every model is trained to convergence on the source lake before adaptation begins.

\stitle{Optimization.} We use Adam with learning rates selected from $\{0.00001, 0.00002, 0.00005\}$ and cross-entropy loss, CTA being multi-class classification. Batch size is $8$ and the maximum BERT sequence length is $128$.

\stitle{\sys hyperparameters.} We set the maximum number of iterations $P=50$, the early-stopping threshold $N_e=5$, and the number of fine-tuning epochs per iteration $N_f=5$. Warm-up set sizes are $50$ tables for P$\to$V and $25$ for V$\to$S, chosen in proportion to target lake size; per-iteration batches $D_{f,l}$ ($l>0$) are $25$ and $15$ columns respectively.

The confidence threshold $\delta$ was selected by sampling $50$ examples from the training set and sweeping $\delta$ from $0$ to $1$, measuring the accuracy of the resulting filtering decisions; $\delta=0.9$ was highest. The mechanism is a two-sided trade-off. With a smaller $\delta$ (\eg $0.8$), fewer columns are forwarded to the verifier, the query set shrinks, and genuine weak samples are missed. With a larger $\delta$ (\eg $0.95$), more columns are forwarded, including ones the annotator already handles correctly, and cost rises for little gain. All hyperparameters are shared across transfers rather than tuned per transfer.

\stitle{Implementations and hardware.} We use the official implementations of all baselines and preserve their settings wherever possible. Sherlock cannot be evaluated at the two lowest-resource V$\to$S budgets ($2.4\%$ and $3.8\%$), because its implementation requires every semantic type to occur at least once in the training set in order to compile the model, and at those budgets the number of training columns is smaller than the type set. We access GPT-3.5-turbo-4k, GPT-4o, and GPT-5.1 through the OpenAI APIs. Experiments ran on Intel Xeon Gold 6240 CPUs with four NVIDIA A800 80GB PCIe GPUs.

\section{Detailed Results}
\label{sec:supp-results}

This section gives the tables and figures behind analyses that the paper reports numerically in prose, so that no number in the paper is unsupported.

\subsection{Reliability of LLM Verification}
\label{sec:supp-verification-acc}

Table~\ref{tab:verification} accompanies \S\ref{sec:reliability-veri}. We prompt ChatGPT on identical columns under the two formulations: direct annotation, where it must select a type from $S_t$, and verification, where it judges the label already assigned by the annotator. Accuracy roughly doubles under the verification formulation on both lakes. Abstentions (``I don't know'') are rare, occurring on $0.009$ and $0.007$ of columns.

As \S\ref{sec:reliability-veri} and \S\ref{sec:limitations} state, this comparison should be read with care. The two formulations have different chance levels, $1/n_t$ against $1/2$, so part of the gap is mechanical. The accuracies are also uncorrected for the base rate of correct labels within the query set: a verifier answering ``Yes'' indiscriminately would score well while carrying no information. The quantities that would settle the matter are that base rate together with the precision and recall of ``No'' decisions against ground truth, since only the ``No'' and ``I don't know'' decisions drive labeling. We do not report them, and accordingly rest the claim that verification decisions carry usable signal on the end-to-end ablation rather than on this table.

\begin{table}[h]
\centering
  \caption{ChatGPT accuracy on identical columns under two prompt formulations.}
  \vspace{-1em}
  \label{tab:verification}
  \small
\begin{tabular}{lcc}
\toprule
                    & Direct annotation & Verification \\
                    \midrule
VizNet     & 0.424                          & 0.807                         \\
Semtab2019 & 0.339                          & 0.870  \\                      \bottomrule
\end{tabular}
\end{table}

\subsection{Effect of Label Set Difference Adjustment}
\label{sec:supp-label}

Table~\ref{tab:cross-knowledge} accompanies \S\ref{sec:label-effect}. We measure the annotator immediately after realignment and warm-up, before any adaptation iteration, so the comparison isolates the transplant. Under \textbf{no-label} the output layer is randomly initialized; under \textbf{label} the columns of shared types are transplanted from the source annotator.

Without the transplant the annotator is essentially inoperative, at or below $0.028$ SW F1 for every backbone on both transfers. With it, \sys(D) and \sys(S) reach $0.241$ and $0.231$ SW F1 on P$\to$V and roughly $0.09$ on V$\to$S. The absolute values are low, as almost no target supervision has been consumed at this point, so these numbers compare initializations rather than end performance. \sys(R) benefits least on P$\to$V ($0.010$), which is consistent with RECA relying on inter-table context that a transplanted output layer alone does not supply.

\begin{table}[h]
  \centering
  \caption{Effect of label set difference adjustment, measured after warm-up and before any adaptation iteration.}
  \vspace{-1em}
  \label{tab:cross-knowledge}
  \small
  \setlength{\tabcolsep}{4pt}
  \begin{tabular}{llcccc}
    \toprule
    &&\multicolumn{2}{c}{no-label}&\multicolumn{2}{c}{label}\\
    \cmidrule(lr){3-4} \cmidrule(lr){5-6}
     & Method & SW F1 & MA F1 & SW F1 & MA F1\\
    \midrule
    \multirow{3}*{P$\to$V} & \sys(D) & 0.028 & 0.004 & 0.241 & 0.075\\
    ~ & \sys(S) & 0.000 & 0.000 & 0.231 & 0.078\\
    ~ & \sys(R) & 0.001 & 0.000 & 0.010 & 0.007 \\
    \midrule
    \multirow{3}*{V$\to$S} & \sys(D) & 0.001 & 0.000 & 0.091 & 0.016 \\
    ~ & \sys(S) & 0.001 & 0.000 & 0.094 & 0.014 \\
    ~ & \sys(R) & 0.000 & 0.000 & 0.090 & 0.013  \\
    \bottomrule
  \end{tabular}
\end{table}

\subsection{Clustering Parameter $K$}
\label{sec:supp-silhouette}

\begin{figure}[h]
\centering
  \subfigure[P$\to$V, SW F1]{\includegraphics[width=0.235\linewidth]{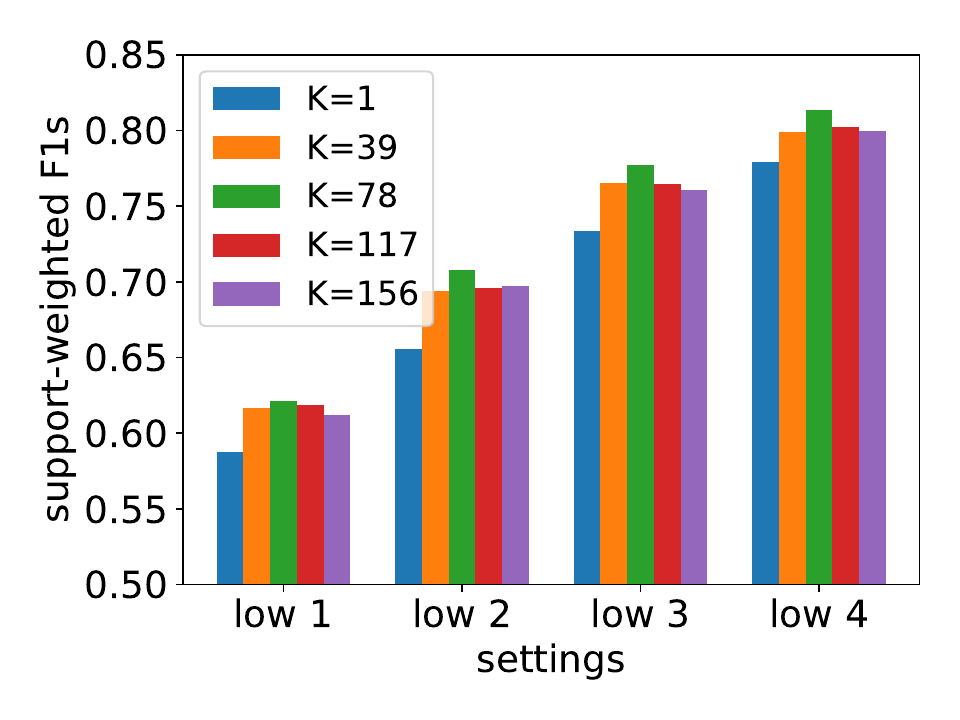}}
  \subfigure[P$\to$V, MA F1]{\includegraphics[width=0.235\linewidth]{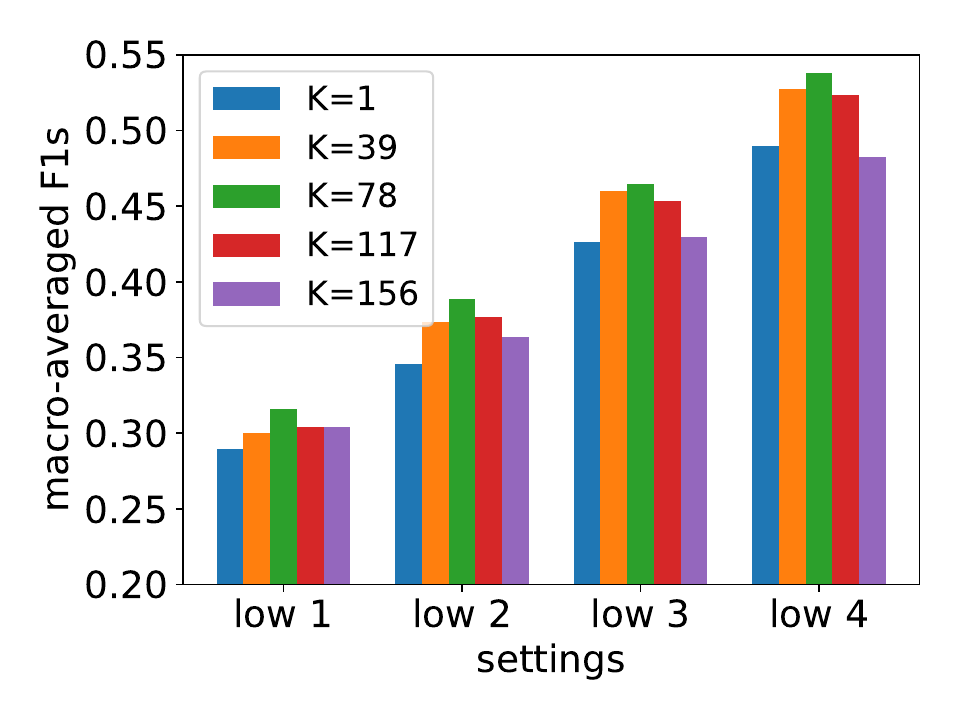}}
  \subfigure[V$\to$S, SW F1]{\includegraphics[width=0.235\linewidth]{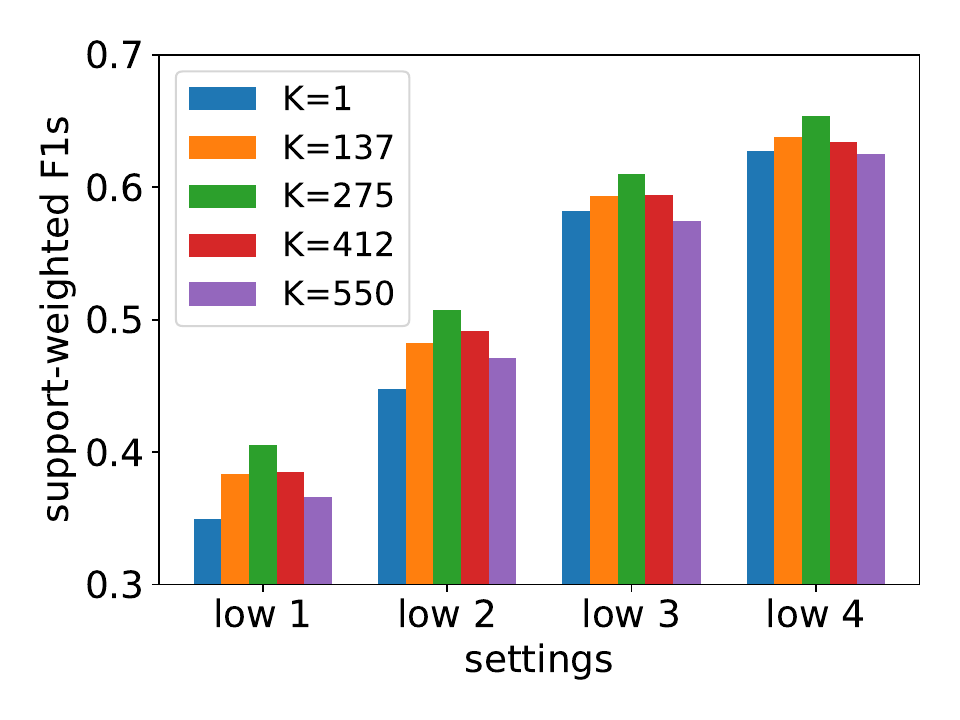}}
  \subfigure[V$\to$S, MA F1]{\includegraphics[width=0.235\linewidth]{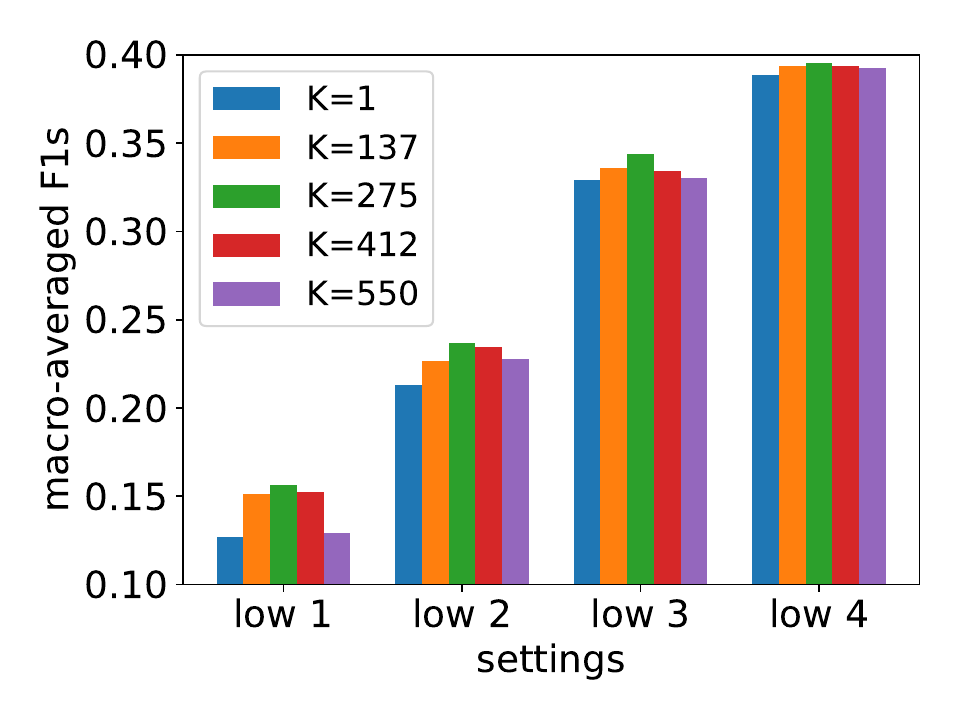}}
  \vspace{-1.2em}
  \caption{Sensitivity to the clustering parameter $K$, averaged over the low-resource budgets.}
  \label{fig:para}
  \vspace{-0.5em}
\end{figure}

Figure~\ref{fig:para} accompanies \S\ref{sec:parameter_sensitivity}. We vary $K$ over $\{1,\; 0.5n_t,\; n_t,\; 1.5n_t,\; 2n_t\}$ and average over the low-resource budgets, using GPT-4o as verifier. Performance peaks at $K=n_t$ on both transfers and in both metrics, matching the rationale of \S\ref{sec:weak-sample}: with roughly $n_t$ type-homogeneous groups in the lake, that granularity propagates a flagged column to columns that share its difficulty without absorbing unrelated ones. Degradation on either side of the peak is gradual, so the choice is not brittle.

As an alternative to fixing $K=n_t$ we selected $K$ by Silhouette score~\cite{rousseeuw1987silhouettes} on Semtab2019, which is highest at $K=255$ against $n_t=275$. Table~\ref{tab:silhouette} compares the two on V$\to$S. The differences are small and inconsistent in direction, $K=255$ being better at the two larger budgets and $K=n_t$ at the two smaller. This is expected: the Silhouette score itself moves only from $0.984$ to $0.987$ between the two settings, so the induced partitions are nearly identical. In practice we suggest choosing $K$ by application: if the extra pass is affordable or the target type distribution is severely imbalanced, compute the Silhouette score; if minimal overhead is the goal, $K=n_t$ is a natural default requiring no extra computation.

\begin{table}[h]
\centering
  \caption{Silhouette-selected $K$ against the default $K=n_t$ on V$\to$S.}
  \vspace{-1em}
  \label{tab:silhouette}
  \small
  \begin{tabular}{lcccc}
    \toprule
    & \multicolumn{2}{c}{Silhouette optimal ($K{=}255$)} & \multicolumn{2}{c}{$n_t$ ($K{=}275$)} \\
    \cmidrule(lr){2-3} \cmidrule(lr){4-5}
    & SW F1 & MA F1 & SW F1 & MA F1 \\
    \midrule
    Low 1 & 0.345 & 0.118 & 0.365 & 0.144\\
    Low 2 & 0.470 & 0.196 & 0.475 & 0.215\\
    Low 3 & 0.579 & 0.314 & 0.573 & 0.296\\
    Low 4 & 0.636 & 0.401 & 0.620 & 0.370\\
  \bottomrule
   \end{tabular}
\end{table}

\subsection{Robustness to Verifier and Encoder}
\label{sec:supp-robust}

Table~\ref{tab:robust} accompanies \S\ref{sec:add-robust}. We replace both ends of the pipeline on P$\to$V: the verifier becomes Llama-3.1-8B-Instruct, an open-weight model far smaller than the hosted alternatives, and the encoder becomes RoBERTa rather than BERT. Compared against \sys(D) in Tables~\ref{tab:sota-viz} and~\ref{tab:sota-full}, gains over the backbone persist at every budget.

The 8B result matters beyond robustness. It shows that the verifier need not be a hosted frontier model, so \sys can be run entirely on premises in deployments where transmitting column values to a third-party API is unacceptable. The Ethical Considerations section discusses this point.

\begin{table}[h]
  \centering
  \caption{Robustness on P$\to$V with an 8B open-weight verifier (Llama-3.1-8B-Instruct) and a RoBERTa encoder.}
  \vspace{-1em}
  \label{tab:robust}
  \small
  \begin{tabular}{lcc}
    \toprule
    Budget & SW F1 & MA F1 \\
    \midrule
    low-1 & 0.718 & 0.370 \\
    low-2 & 0.754 & 0.453 \\
    low-3 & 0.807 & 0.516 \\
    low-4 & 0.827 & 0.514 \\
    25\% & 0.858 & 0.610 \\
    50\% & 0.882 & 0.694 \\
    100\% & 0.919 & 0.789 \\
    \bottomrule
  \end{tabular}
\end{table}